\documentclass[10pt,twocolumn,letterpaper]{article}

\usepackage[pagenumbers]{cvpr} 

\usepackage{multirow}
\usepackage{algorithmic}
\usepackage{algorithm}
\usepackage{array}
\usepackage{textcomp}
\usepackage{stfloats}
\usepackage{url}
\usepackage{verbatim}

\usepackage{cite}

\usepackage{multirow}
\usepackage{makecell}   
\usepackage{amsfonts}   
\usepackage{amsmath}    
\usepackage{cite}      
\usepackage{booktabs}   
\usepackage{bm}         
\usepackage{setspace}
\usepackage{tikz}
\usepackage{pgfplots}
\usepackage{amssymb}
\usepackage{adjustbox}
\usepackage{float}
\usepackage{subcaption}
\usepackage{graphicx}

\definecolor{cvprblue}{rgb}{0.21,0.49,0.74}
\usepackage[pagebackref,breaklinks,colorlinks,citecolor=cvprblue]{hyperref}

\def\paperID{10559} 
\def\confName{CVPR}
\def\confYear{2024}

\title{LoopSparseGS: Loop Based Sparse-View Friendly Gaussian Splatting}

\author{
Zhenyu Bao\footnotemark[1] \textsuperscript{ 1,2}, 
Guibiao Liao\footnotemark[2] \textsuperscript{ 1,2},
Kaichen Zhou\textsuperscript{1},
Kanglin Liu\textsuperscript{2}, 
Qing Li\footnotemark[2] \textsuperscript{ 2},
Guoping Qiu\textsuperscript{3} \vspace{0.1 em} \and 
$^{1}$Peking University  \and  $^{2}$Pengcheng Laboratory \and  $^{3}$University of Nottingham 
}

\begin{document}

\twocolumn[{%
	\renewcommand\twocolumn[1][]{#1}%
        \vspace{-1 em}
	\maketitle
	\includegraphics[width=\linewidth]{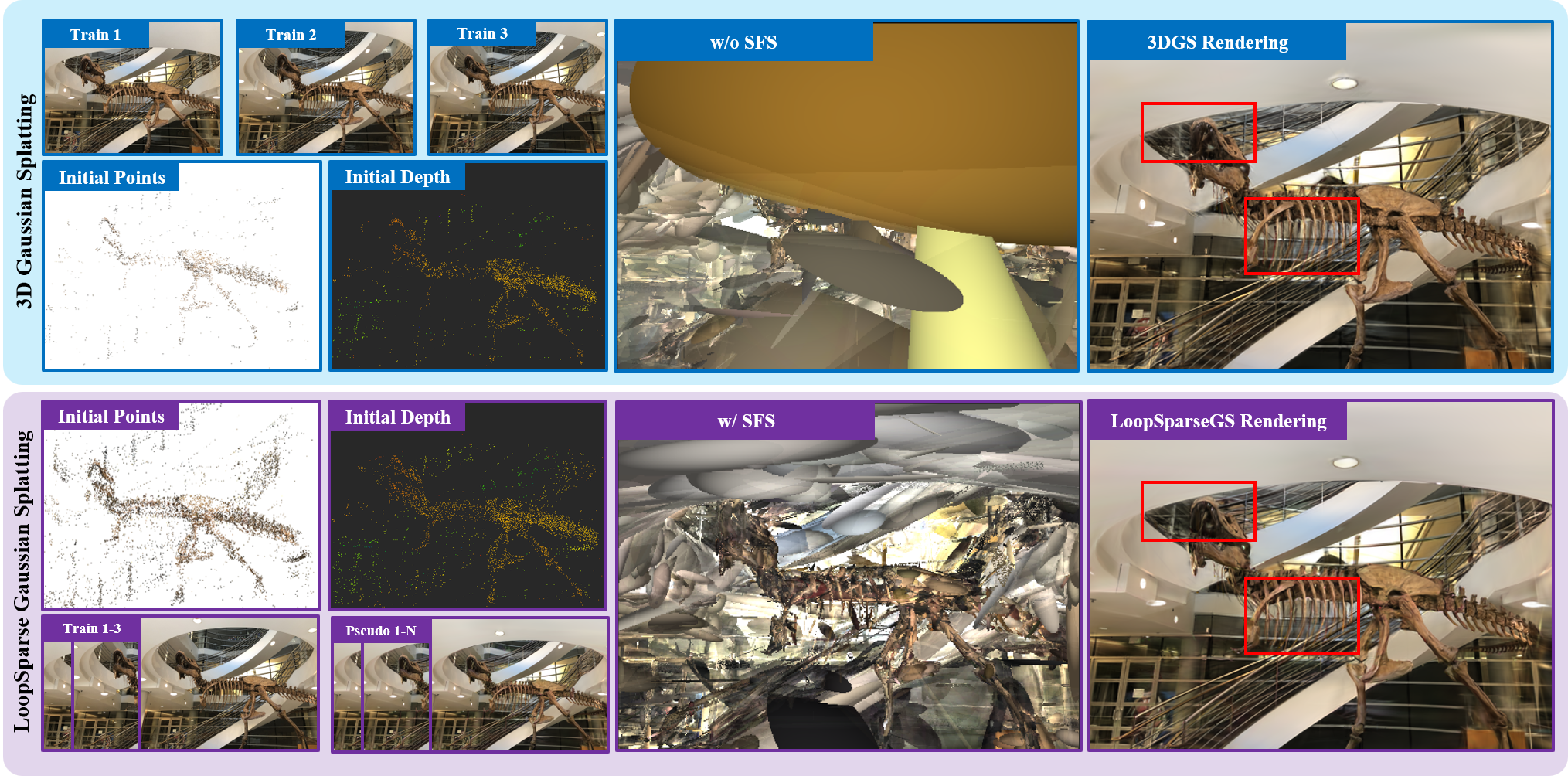}
	\captionof{figure}{In scenarios with limited input data, the standard 3D Gaussian Splatting (3DGS) method generates insufficient points and minimal depth constraints for training. Our LoopSparseGS employs additional pseudo-cameras to produce more comprehensive initialization points and richer depth information for 3DGS training. Additionally, we found that the excessively large ellipsoids, damage view rendering quality through blurring. To mitigate the issue, we propose a Sparse Friendly Sampling (SFS) strategy to split oversized ellipsoids. The results are presented in the third column, demonstrating the effectiveness of our method.}
        \vspace{2em}
	\label{fig:teaser_figure}
}]
\maketitle
\renewcommand{\thefootnote}{\fnsymbol{footnote}}
\footnotetext[1]{$ $ Github: \href{https://github.com/pcl3dv/LoopSparseGS}{\textcolor{magenta}{https://github.com/pcl3dv/LoopSparseGS}}}
\footnotetext[2]{$ $ Corresponding authors}

\begin{abstract}
Despite the photorealistic novel view synthesis (NVS) performance achieved by the original 3D Gaussian splatting (3DGS), its rendering quality significantly degrades with sparse input views. This performance drop is mainly caused by the limited number of initial points generated from the sparse input, insufficient supervision during the training process, and inadequate regularization of the oversized Gaussian ellipsoids. To handle these issues, we propose the LoopSparseGS, a loop-based 3DGS framework for the sparse novel view synthesis task. In specific, we propose a loop-based Progressive Gaussian Initialization (PGI) strategy that could iteratively densify the initialized point cloud using the rendered pseudo images during the training process. Then, the sparse and reliable depth from the Structure from Motion, and the window-based dense monocular depth are leveraged to provide precise geometric supervision via the proposed Depth-alignment Regularization (DAR). Additionally, we introduce a novel Sparse-friendly Sampling (SFS) strategy to handle oversized Gaussian ellipsoids leading to large pixel errors. Comprehensive experiments on four datasets demonstrate that LoopSparseGS outperforms existing state-of-the-art methods for sparse-input novel view synthesis, across indoor, outdoor, and object-level scenes with various image resolutions. 
\end{abstract}
\section{Introduction}
\label{sec:intro}

Novel view synthesis (NVS) aims to generate photorealistic images of 3D scenes from perspectives that were not originally captured \cite{mildenhall2020nerf,zhang2020nerf++,nerfreview,bao20243d,instantngp,liao2024clip,scaffoldgs,zhou2024dynpoint}, which is an essential task in computer vision and graphics field. Recently, 3D Gaussian Splatting (3DGS) \cite{3dgs} has emerged as a promising technique for NVS, as it can efficiently model the highly detailed appearance and geometry of 3D scenes.
Such superior performance is usually obtained when large amounts of input images are available. 
However, in many real-world applications \cite{robotnavigation,liao2024vlm2scene,conceptfusion,zhou2024neural}, only a few sparse input images are available such as in sports event broadcasting and robotics, where acquiring dense views is often time-consuming and expensive, even impossible. 
These sparse inputs introduce several challenges to 3DGS. 
Firstly, given the sparse input views, the initial Gaussian points provided by Structure from Motion (SfM) \cite{schonberger2016structure} can be sparse and inadequate, as shown in Fig. \ref{fig:teaser_figure} (top left). Secondly, reconstructing the appearance and geometry of scenes becomes an under-constrained and ill-posed issue with insufficient inputs with only the image reconstruction constraints. 
Thirdly, the scales of some Gaussians grow to be very large during the optimization process, and these oversized Gaussian ellipsoids result in the overfitting problem, thus producing unsatisfactory results at novel viewpoints as illustrated in Fig. 1 (top middle).. 


Recent studies \cite{chung2024depth,zhu2023fsgs,li2024dngaussian} have attempted to address the aforementioned issues of sparse-input 3DGS. 
To handle the issue of sparse initialized Gaussian points, FSGS \cite{zhu2023fsgs} introduces a proximity-guided Gaussian unpooling technique, which generates new Gaussians by measuring the proximity of existing Gaussians with their neighbours during training. This densification strategy, however, is sensitive to noisy points, potentially leading to the generation of invalid points. 
To mitigate the under-constrained problem, DNGaussian \cite{li2024dngaussian} adopts monocular depth, obtained from the pre-trained monocular depth estimator \cite{MiDaS}, to constrain the depth rendered by 3DGS in a global-local normalization manner. However, these monocular depth maps are often scale-inconsistent across different views, posing a challenge for effective depth regularization. 
Additionally, previous studies ignore the issue of excessively large Gaussians in sparse-input scenarios, limiting the quality of novel view synthesis.

In this paper, we present the \textbf{LoopSparseGS}, a novel 3DGS framework for precise and robust sparse-input novel view synthesis. LoopSparseGS is built upon a looping mechanism and incorporates a sparse-friendly Gaussian densification strategy with the following considerations.

As shown in Fig. \ref{fig:teaser_figure} (bottom left), we observe that those rendered views close to the training views exhibit high visual quality even with sparse input. 
This observation motivates us to integrate these pseudo images, i.e., rendered novel images, with training images to generate additional initialized 3D points using SfM. 
This process is developed in a looping mechanism to increase the number of initialized points. 
Consequently, we propose a \textbf{Progressive Gaussian Initialization (PGI)} strategy, which leverages both rendered images and training images to iteratively increase initialized Gaussian points, resulting in more comprehensive scene coverage. 

Moreover, increased initialized 3D points provide additional precise but sparse depth constraints for 3DGS optimization. Dense monocular depth from the pre-trained model provides dense but scale-invariant depth constraints. To effectively utilize the two constraints, we develop a \textbf{Depth-alignment Regularization (DAR)} approach to generate smoother and more precise rendered depths as illustrated in (d) of Fig. \ref{fig:depth_align}. 


Furthermore, to address the issue of excessively large Gaussian ellipsoids, we propose a \textbf{Sparse-friendly Sampling (SFS)} strategy guided by the pixel error. Specifically, SFS identifies and splits the Gaussian ellipsoids associated with high-error pixels that have the largest weights, which could
effectively produce more detailed geometric and rendering results, as shown in Fig. \ref{fig:teaser_figure} (bottom middle). 
The main contributions of this work are as follows: 
\begin{itemize}
    \item We present the LoopSparseGS, a novel 3DGS-based framework for sparse-input novel view synthesis, featuring a looping mechanism to provide denser Gaussian initialization and precise geometry constraints, and a sparse-friendly sampling strategy to address the oversized Gaussian ellipsoids. 
    \item We develop a Progressive Gaussian Initialization (PGI) method to produce dense 3D Gaussian points by incorporating iteratively rendered images with training images into SfM.
    \item We propose a Depth Alignment Regularization (DAR) approach that aligns dense relative-scale monocular depths with absolute-scale sparse SfM-derived depths, to provide effective geometric constraints.
    \item We introduce a Sparse-friendly Sampling (SFS) strategy to address the issue of excessively large Gaussian ellipsoids unique to sparse-input scenes, thus further enhancing the view synthesis quality of the scene. 
    \item Comprehensive experimental results on four datasets demonstrate that our proposed approach outperforms existing state-of-the-art methods in novel view synthesis with sparse-input data, across indoor scenes, outdoor scenes, and object-level scenes.    
\end{itemize}
\section{Related Work}
\subsection{Novel View Synthesis using Radiance Fields} 
Novel view synthesis techniques typically utilize one or more input views to generate images from novel perspectives. Recent advancements in this field have concentrated on employing radiance fields and achieved encouraging progress.
For example, Mildenhall et al. \cite{mildenhall2020nerf} introduce Neural Radiance Field (NeRF) that enables novel view synthesis using coordinate-based neural networks. 
Tremendous following efforts concentrate on improving its rendering quality \cite{barron2021mip,barron2022mip,barron2023zip}, efficiency \cite{chen2022tensorf,fridovich2022plenoxels,muller2022instant,zhang2023fast}, scene understanding \cite{zhi2021place,kerr2023lerf,liao2024ov,zheng2024surface}, and 3D generation \cite{gu2021stylenerf,hollein2023text2room}. 
Particularly, Mip-NeRF \cite{barron2021mip} employs conical frustum instead of single rays to reduce aliasing. Mip-NeRF 360 \cite{barron2022mip} extends this approach to handle unbounded scenes.  
Recently, Kerbl et al.\cite{3dgs} achieve a significant breakthrough with the 3D Gaussian Splatting (3DGS) method, which enhances rendering efficiency using explicit Gaussian representations and the differentiable rasterization technique. 
Building upon its high efficiency in novel view synthesis, several works attempt to extend 3DGS to various tasks. Wu et al. \cite{wu20244d} propose an explicit representation method for dynamic scenes utilizing 3D Gaussian and 4D neural voxels. Tang et al. \cite{tang2023dreamgaussian} presents a generative 3D Gaussian Splatting model for efficient text-to-3D content creation. Zhou et al. \cite{zhou2024drivinggaussian} introduce DrivingGaussian for efficient dynamic autonomous driving scene reconstruction. 

Although the methods mentioned above demonstrate excellent performance in novel view synthesis, they typically require dense input views for training the radiance fields. When provided with sparse training views, these methods tend to overfit the available training views, resulting in a significant performance drop in novel views.

\subsection{Sparse Novel View Synthesis}
In recent years, several NeRF-based studies have been proposed to address sparse-input novel view synthesis. 
Specifically, RegNeRF \cite{niemeyer2022regnerf} introduces geometry and color regularization from unobserved viewpoints, enhancing the quality of sparse-input novel view synthesis. It employs a 2D consistency loss on the depth and color of image patches, ensuring that neighboring regions have similar geometry and appearance. 
InfoNeRF \cite{kim2022infonerf} enhances sparse-input view synthesis by employing regularization techniques based on information theory. Specifically, it applies a sparsity constraint on the density distribution of the ray by minimizing entropy.
DS-NeRF \cite{deng2022depth} utilizes sparse depth cues generated by SfM, to impose depth supervision for sparse NeRF.  
ViP-NeRF \cite{somraj2023vip} enhances the traditional NeRF framework by incorporating the visibility prior, which enforces multi-view constraints during optimization. This modification involves calculating the visibility of a point and using these results to regularize the visibility and alpha-blended depth across different views. 
FreeNeRF \cite{yang2023freenerf} incorporates a frequency regularization strategy designed to train the sparse-input NeRF, aiming to regularize the frequency range of NeRF’s inputs, and the other to penalize the near-camera density fields. 
SparseNeRF \cite{wang2023sparsenerf} utilizes a pre-trained depth estimation model to generate pseudo-ground truth depth maps, which are employed for a local depth ranking loss. Besides, SparseNeRF applies a depth smoothness loss to ensure that the rendered depth maps exhibit patch-wise smoothness.

In addition, some 3DGS-based approaches try to tackle the sparse-input novel view synthesis. 
For example, Yu et al. \cite{chung2024depth} align sparse depth from SfM with dense depth from a monocular depth estimation model \cite{MiDaS} to guide the geometry for sparse-input 3DGS. 
FSGS \cite{zhu2023fsgs} integrates estimated monocular depth and employs a Pearson correlation depth distribution loss to train sparse 3D Gaussian Splatting. 
Likewise, DNGaussian \cite{li2024dngaussian} introduces a monocular depth loss and incorporates global-local depth normalization to optimize the parameters of Gaussians. 
SparseGS \cite{xiong2023sparsegs} proposes generative constraints from a pre-trained diffusion model \cite{rombach2022high}, which guides the 3D Gaussian representation in novel views via Score Distillation Sampling.

Unlike previous methods that utilize scale-inconsistent monocular depth across different views for regularization, our work introduces a Depth-alignment Regularization (DAR) approach. DAR extract accurate and reliable depth values from the SfM points and aligns them with monocular depths using a sliding-window mechanism, providing more effective geometric supervision. 
Furthermore, our work proposes a loop-based Gaussian initialization, resulting in a denser point cloud. This not only offers more precise depth values for the DAR but also facilitates the training convergence quality of Gaussians. 
Additionally, we present a sparse-friendly sampling strategy to further enhance Gaussian densification.



\begin{figure*}[t]
  \centering
  \includegraphics[width=\linewidth]{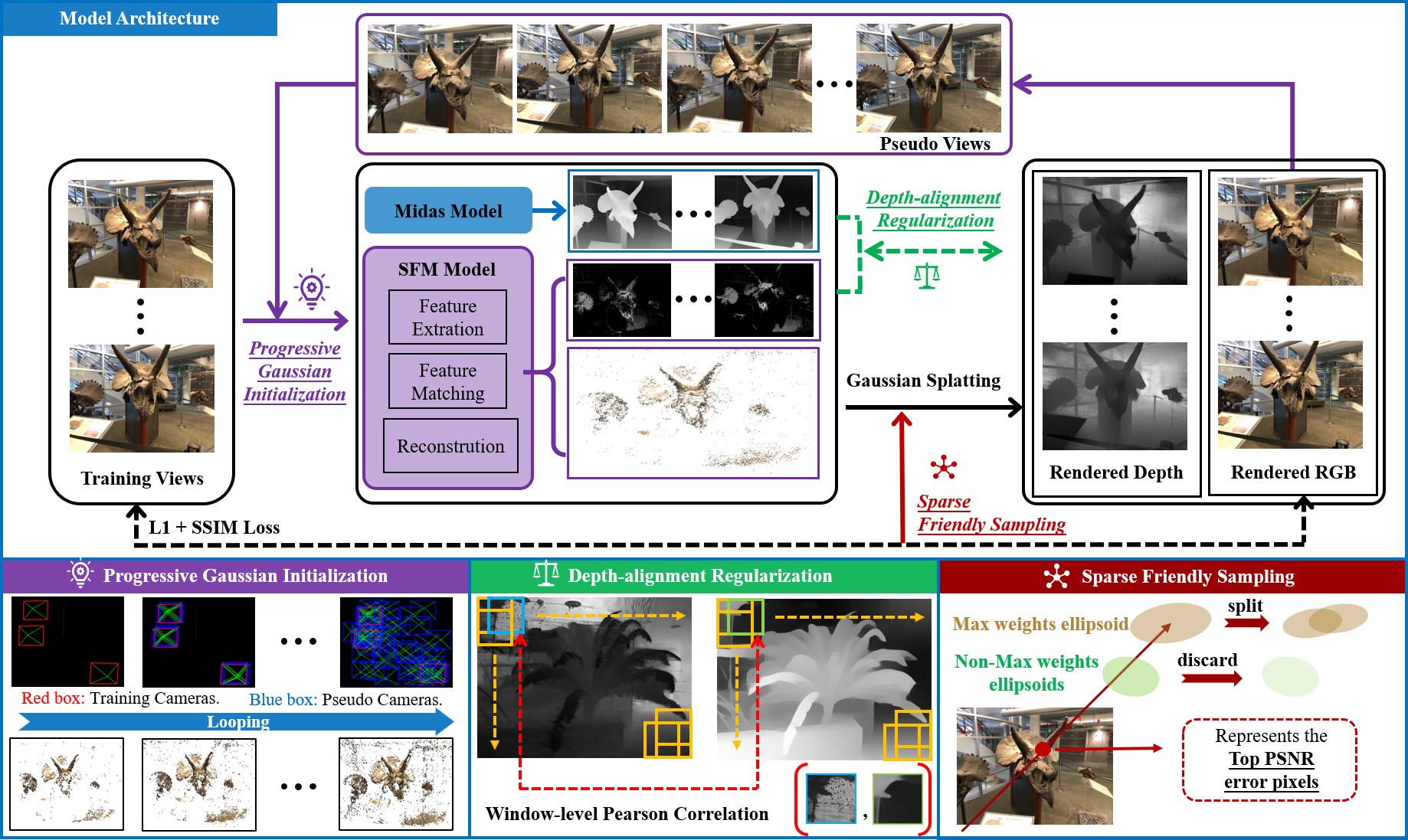}
   \caption{Overview of the proposed \textbf{LoopSparseGS}.
   The LoopSparseGS features three key components: Progressive Gaussian Initialization, Depth Alignment Regulerizer and Sparse-friendly sampling. Progressive Gaussian Initialization leverages the training view and high-quality pseudo views near the training view to increase the number of the Gaussian initialized points. Depth Alignment Regularizer incorporates the precise SFM depth and monocular depth and provides a sliding window-based manner to align the two scale-invariant depth regularizers. Sparse-friendly sampling slit large Gaussian ellipsoids of large pixels errors to enhance the representation capacity of large pixel areas. 
   }
   \label{fig:pipeline}
\end{figure*}
\section{Method}
\label{sec:method}
LoopSparseGS facilitates scene novel view synthesis given sparse input images, and the framework is illustrated in Fig. \ref{fig:pipeline}. 
First, initialized sparse point clouds and camera parameters are obtained using Structure from Motion (SfM). We then introduce a loop-based initialization strategy using pseudo-view rendering results to progressively provide denser Gaussian initialization (Section \ref{method_PGI}). 
Second, during the Gaussian optimization, we incorporate depth-alignment regularization to impose additional and precise geometric constraints (Section \ref{method_DAR}). 
Lastly, we adopt a sparse-friendly Gaussian densification approach to sample effective Gaussians for sparse-input reconstruction quality enhancement (Section \ref{method_SFS}). 
Before introducing our method, we briefly revisit 3D Gaussian Splatting in Section \ref{Preliminaries}.

\subsection{Preliminaries} \label{Preliminaries} 
3D Gaussian Splatting (3DGS) \cite{3dgs} represents a 3D scene using a set of anisotropic 3D Gaussian primitives, enabling efficient and differentiable rendering via $\alpha$-blending.   
The properties of $i$-th Gaussian primitive can be described as ${\Theta}_{i} = \{ u_{i}, {o}_{i}, s_i, q_i, c_i \}$, where $u_{i} \in \mathbb{R}^3$ is the center, ${o}_{i} \in \mathbb{R}$ is the opacity, $s_i \in \mathbb{R}^3$ is the scaling factor, $q_i \in \mathbb{R}^4$ is the rotation, and $c_i \in \mathbb{R}^3$ is the color.  
To compute the pixel color $C$, 3DGS employs differentiable $\alpha$-blending point-based rendering by blending $\mathcal{N}$ Gaussian points in the front-to-back depth order, which can be written as: 
\begin{equation}
C = \sum_{i \in \mathcal{N}} \mathbf{c}_i \alpha_i \prod_{j=1}^{i-1} (1 - \alpha_j), 
\end{equation} 
where \(\mathcal{N}\) denotes the set of Gaussian points that overlap with the given pixel, and \(\alpha_{i}\) is calculated by \(\alpha_{i} = o_i f_i^{2D}\), where \(f_i^{2D}\) represents the projection function of the \(i\)-th Gaussian onto the 2D plane. 

3DGS is optimized by projecting 3D Gaussians onto the 2D image plane and employing gradient-based color supervision to minimize the distance between the rendered image $\tilde{I}$ and the ground truth image ${I}$. This process is as follows: 
\begin{equation}
\mathcal{L}_{color} = (1-\lambda)\mathcal{L}_{1}(\tilde{I}, I) + \lambda\mathcal{L}_{D-SSIM}(\tilde{I}, I), 
\end{equation}
where $\lambda$ is set to 0.2 as per \cite{3dgs}.

\subsection{Progressive Gaussian Initialization (PGI)} \label{method_PGI} 

Initialization of Gaussian points is crucial for 3DGS-based novel view synthesis, as it significantly impacts the training convergence quality and speed. The original 3DGS relies on dense input images to generate enough initial Gaussian points. However, in scenarios with limited views, the number of 3D points drops dramatically, potentially compromising reconstruction quality. 
Considering that rendered views close to the training views exhibit satisfactory visual quality, as illustrated in Fig. \ref{fig:teaser_figure} (bottom left), we develop a \textbf{Progressive Gaussian Initialization (PGI)} approach, which combines the rendered images with the original training images to generate additional initialized points. Instead of generating images once, we rely on the iteratively refined 3DGS to produce the high-quality pseudo images progressively. The detailed process is shown in Fig.\ref{fig:pipeline}. 

Before starting a new loop,  we generate 4 new pseudo-views around each training view, resulting in a total of $P \times 4$ new pseudo-views per loop iteration, where $P$ denotes the number of training views. 
The camera location of the pseudo-views is obtained by adding Gaussian noise to the training camera positions. 
The locations of the generated pseudo views are computed as follows:

\begin{equation}
      z_{gl}^{ij} = N(z_{tj}\ , \ \varepsilon +\delta \times l)^{i}, j\in [1,P], i\in [1,4]
      \label{eq:pseudo camera}
\end{equation}
Where $z_g$ and $z_t$ are locations of the pseudo views and training views, respectively. $N$ is Gaussian noise, $j$ denotes the $j-$th training view, and $i$ denotes the $i-$th pseudo-view generated from one of the training views for each loop. 
To ensure the quality of the generated pseudo-view, $\varepsilon$ starts from a small value. It gradually enlarges by a rate of $\delta$ with loop iterating ($l$). 
$\varepsilon$ and $\delta$ are set to $0.02$ and $0.1$, respectively. 
As the number of loops increases, the pseudo view gradually expands the coverage around the training views.
To force the view range of the pseudo-images under the coverage of the training views, the locations of pseudo-images should not be out of the bounding box determined by the locations of the training views. The orientation of pseudo-images is the averages of two adjacent training views as in \cite{zhu2023fsgs}.
For each loop, pseudo-images generated from previous loops are accumulated to train the Gaussians. 


\subsection{Depth-alignment Regularization (DAR)} \label{method_DAR} 

\begin{figure}[t!]   
	\centering
	\begin{subfigure}{\linewidth}
            \rotatebox[origin=c]{90}{\footnotesize{w/o Filter}\hspace{-1.3cm}}
            \begin{minipage}[t]{0.235\linewidth}
                \centering
                \includegraphics[width=1\linewidth]{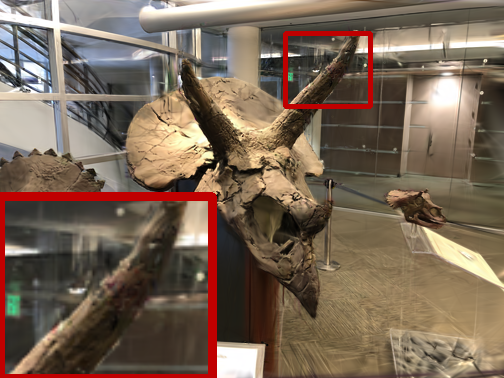}
            \end{minipage}
            \begin{minipage}[t]{0.235\linewidth}
                \centering
                \includegraphics[width=1\linewidth]{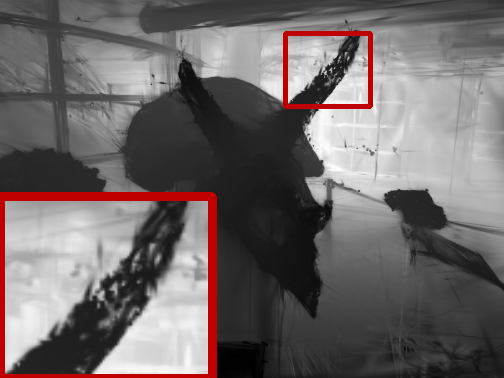}
            \end{minipage}
            \begin{minipage}[t]{0.235\linewidth}
                \centering
                \includegraphics[width=1\linewidth]{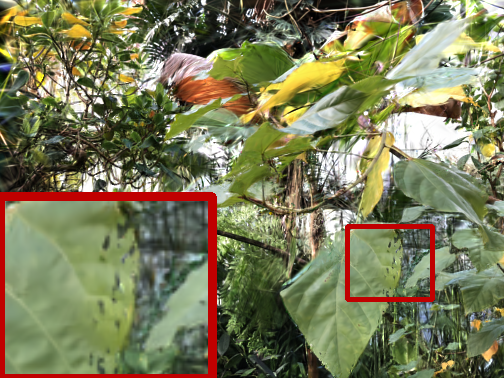}
            \end{minipage}
            \begin{minipage}[t]{0.235\linewidth}
                \centering
                \includegraphics[width=1\linewidth]{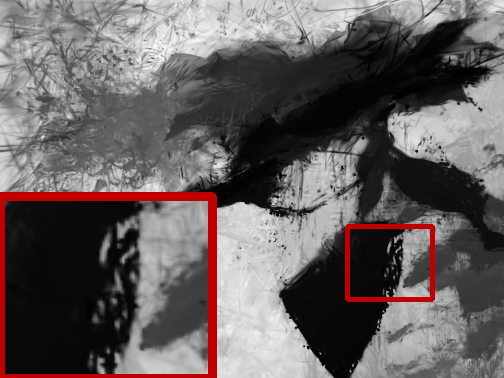}
            \end{minipage}
        \end{subfigure}

        \vspace{0.1cm}
	\begin{subfigure}{\linewidth}
            \rotatebox[origin=c]{90}{\footnotesize{w/ Filter}\hspace{-1.3cm}}
            \begin{minipage}[t]{0.235\linewidth}
                \centering
                \includegraphics[width=1\linewidth]{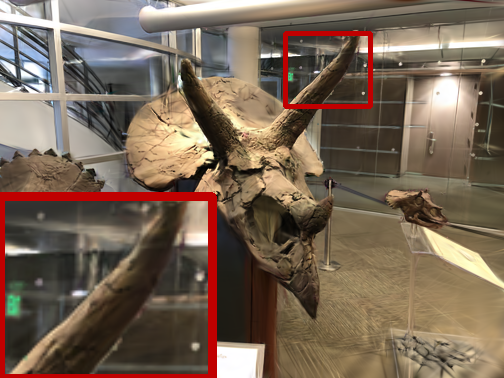}
                \caption{}
            \end{minipage}
            \begin{minipage}[t]{0.235\linewidth}
                \centering
                \includegraphics[width=1\linewidth]{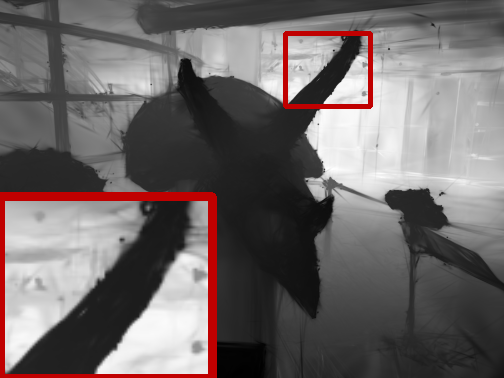}
                \caption{}
            \end{minipage}
            \begin{minipage}[t]{0.235\linewidth}
                \centering
                \includegraphics[width=1\linewidth]{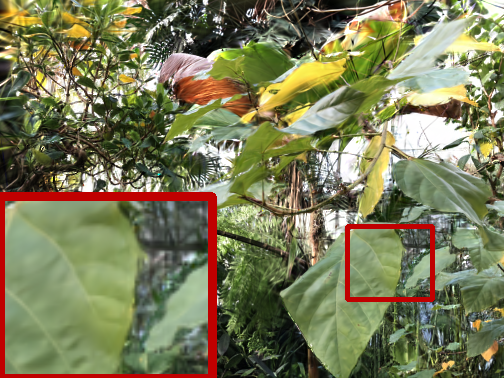}
                \caption{}
            \end{minipage}
            \begin{minipage}[t]{0.235\linewidth}
                \centering
                \includegraphics[width=1\linewidth]{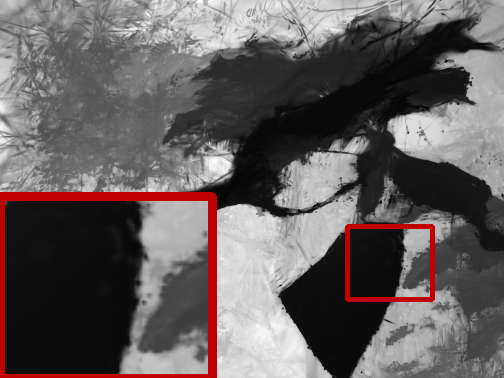}
                \caption{}
            \end{minipage}
        \end{subfigure}

       \caption{Illustration of the rendered RGB and depth maps without using filter strategy (``w/o" Filter) and utilizing filter strategy (``w" Filter). (a) Rendered image of \textit{Horns}. (b) Rendered depth of \textit{Horns}. (c) Rendered image of \textit{leaves}.  (d) Rendered depth of \textit{leaves}. Without filtering, the rendered depth shows significant holes in the edges of the horn and leaves, while the holes are filled up when using our custom filter strategy.}
       \label{fig:depth_mask}   
\end{figure}


Beyond providing denser initialized Gaussian points, our loop-based method offers an additional advantage: valuable and reliable depth information derived from these initialized 3D points. 
Given the inherent inaccuracies and sparsity in the depth supervision from these 3D points, we propose a Depth-alignment Regularization (DAR) strategy, which comprises a Filter Enhancement and a Sliding window-based Alignment.
The former aims to enhance the accuracy of SfM-derived depth while the latter incorporates dense monocular depth cues to improve depth regularization for sparse-input 3D Gaussian optimization.

\textbf{Filter Enhancement.}
We observed that inaccurate depth information directly derived from SfM can result in erroneous rendering, as illustrated in Fig. \ref{fig:depth_mask}. The \textit{horns} and \textit{leaves} scenes exhibit significant holes at their edges. To enhance depth reliability, we implemented three filtering strategies according to the reliability and visibility of matched points from certain perspectives. 

\textit{Filter Strategy 1}: Firstly, we ignore the depth of 3D points with large match errors according to the SfM key point match report. We use the average pixel match error to find coarse points. The threshold is set to 2. This filter strategy can be described with the following equation.

\begin{equation}
    D(p)=\begin{cases}
        d,\ \text{if} \ \Upsilon (p)<2,
        \\
        0,\ \text{if} \ \Upsilon (p)>2,
    \end{cases}
\end{equation}
where $\Upsilon$ is the average pixel match error and $d$ is the depth value. Note that such points are not used for depth map generation but are yet kept for Gaussian initialization. Although such coarse points can not produce precise depth information they are accurate enough to initialize Gaussian.

\textit{Filter Strategy 2}: Secondly, considering that pseudo-view images may coincidentally have erroneous regions that satisfy the key points match, we discard points generated solely from pseudo-view. This is illustrated as follows:
\begin{equation}
      p=
      \begin{cases} 
            \text{Keep}, \ \text{if} \ p \subset 
            M(C^{t}_{i},C^{t}_{j}) || M(C^{t}_{i},C^{g}_{j}),
            \\ 
            \text{Discard}, \ \text{if} \ p \subset    
            M(C^{g}_{i},C^{g}_{j}),
      \end{cases}
      \label{eq:filter1}
\end{equation}
where $p$ is the matched 3D point. $C^{t}$ and $C^{g}$ are the training views and pseudo views, respectively. The $M$ indicates that the point is computed from these views. Points are unreliable when they are produced from pseudo-images only rendered by the preliminary trained 3DGS.



\textit{Filter Strategy 3}: Lastly, considering that the foreground points block the background points in 3D space, we select 3D points to produce the depth images based on visibility provided by the RGB images.
For certain view $C_{i}$, its corresponding depth $D_{i}$ of certain 3D point $p_{j}$ is $d$, only if $p_{j}$ derived from view $C_{i}$. This can be described under the following conditions:
\begin{equation}
      D_{i}(p_{j})=\begin{cases}
            d,\ \text{if} \ p_{j} \subset M(C_{i},C_{k}),
            \\
            0,\ \text{other situations},
       \end{cases}
      \label{eq:filter2}
\end{equation}
where $ D_{i}(p_{j})$ represents that the depth of point $p_{j}$ in $i$-th camera. $C_i$ is the $i$-th camera. $d$ represents the distance from point $p_{j}$ to camera $C_i$.

\begin{figure}[t!]  
	\centering
	\begin{subfigure}{\linewidth}
            \rotatebox[origin=c]{90}{\footnotesize{Fern}\hspace{-1.3cm}}
            \begin{minipage}[t]{0.235\linewidth}
                \centering
                \includegraphics[width=1\linewidth]{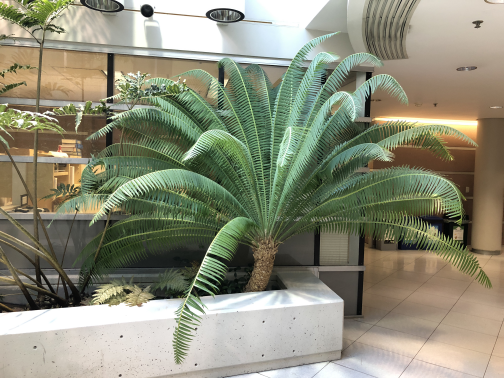}
            \end{minipage}
            \begin{minipage}[t]{0.235\linewidth}
                \centering
                \includegraphics[width=1\linewidth]{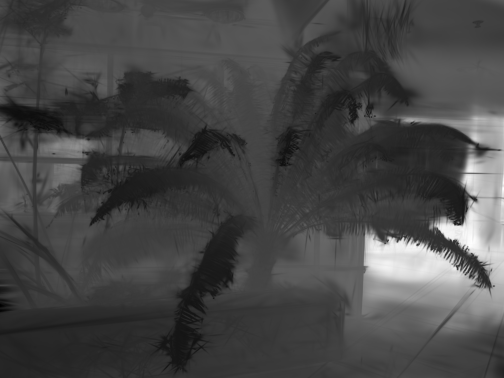}
            \end{minipage}
            \begin{minipage}[t]{0.235\linewidth}
                \centering
                \includegraphics[width=1\linewidth]{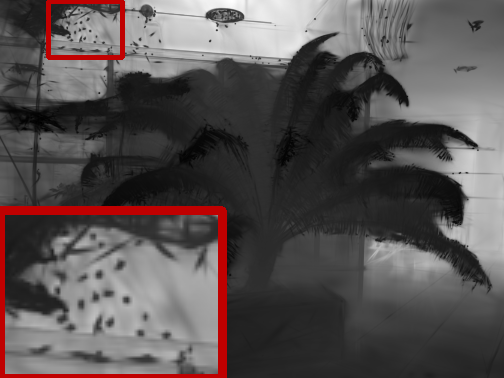}
            \end{minipage}
            \begin{minipage}[t]{0.235\linewidth}
                \centering
                \includegraphics[width=1\linewidth]{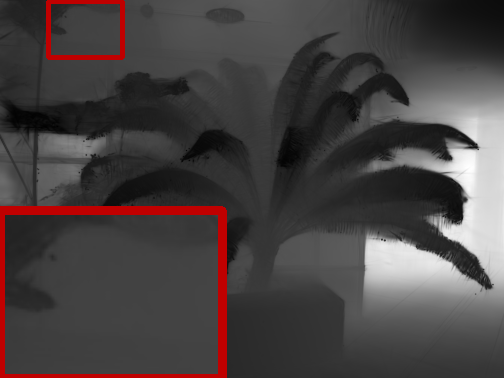}
            \end{minipage}
        \end{subfigure}

        \vspace{0.1cm}
	\begin{subfigure}{\linewidth}
            \rotatebox[origin=c]{90}{\footnotesize{Trex}\hspace{-1.3cm}}
            \begin{minipage}[t]{0.235\linewidth}
                \centering
                \includegraphics[width=1\linewidth]{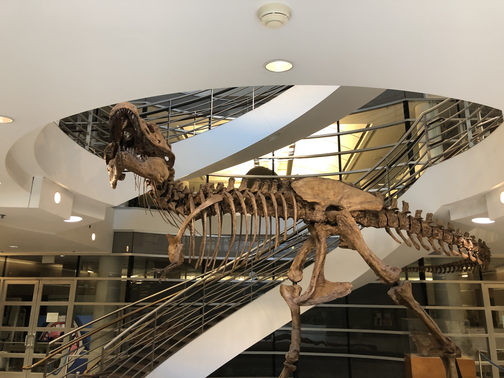}
                \caption{}
            \end{minipage}
            \begin{minipage}[t]{0.235\linewidth}
                \centering
                \includegraphics[width=1\linewidth]{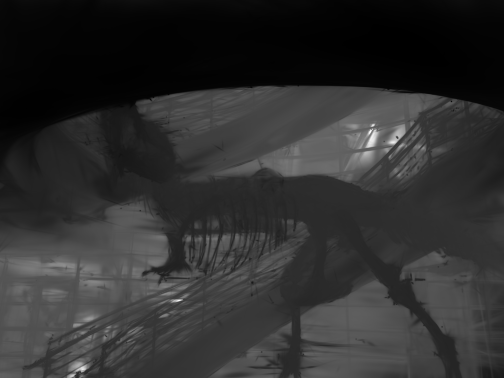}
                \caption{}
            \end{minipage}
            \begin{minipage}[t]{0.235\linewidth}
                \centering
                \includegraphics[width=1\linewidth]{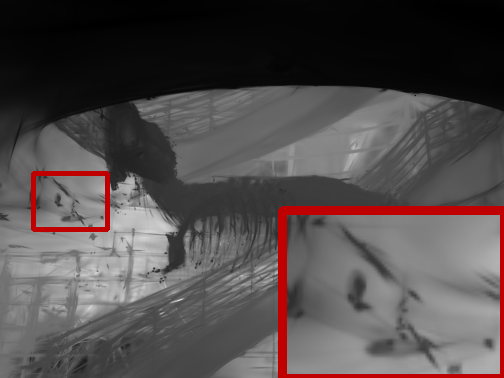}
                \caption{}
            \end{minipage}
            \begin{minipage}[t]{0.235\linewidth}
                \centering
                \includegraphics[width=1\linewidth]{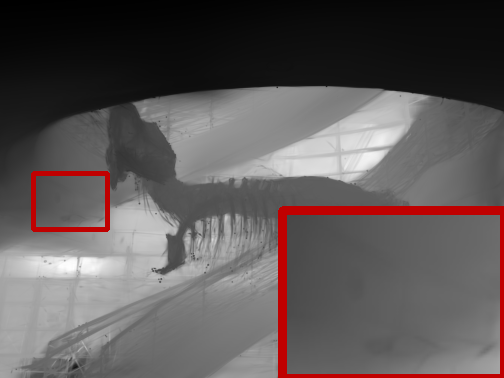}
                \caption{}
            \end{minipage}
        \end{subfigure}

       \caption{Illustration of the rendered depth maps using different depth supervision. (a) GT image. (b) Using SfM-derived depth supervision. (c) Using SfM-derived depth and Monocular depth supervision without depth alignment. (d) Using SfM-derived depth and Monocular depth with our depth-alignment strategy. }
       \label{fig:depth_align}
\end{figure}

\textbf{Sliding window-based Alignment.}
To impose dense depth regularization, an intuitive approach is to employ the image-level \textit{Pearson} distribution loss between the rendered depth and the monocular depth predicted by the MiDaS model \cite{MiDaS}. 
However, directly combining SfM-derived sparse depth and dense monocular depth constraints leads to a misalignment issue, manifesting as erroneous black points in the rendered depth maps, as illustrated in (c) of Fig. \ref{fig:depth_align}. 

This phenomenon is attributed to the limitations of the relative-scale and image-level Pearson constraints, which struggle to impose sufficient constraints on local regions, thereby hindering the alignment of absolute-scale and sparse depths derived from SfM.

\begin{figure}[t!]  
	\centering
	\begin{subfigure}{\linewidth}
            \begin{minipage}[t]{0.41\linewidth}
                \centering
                \includegraphics[width=1\linewidth]{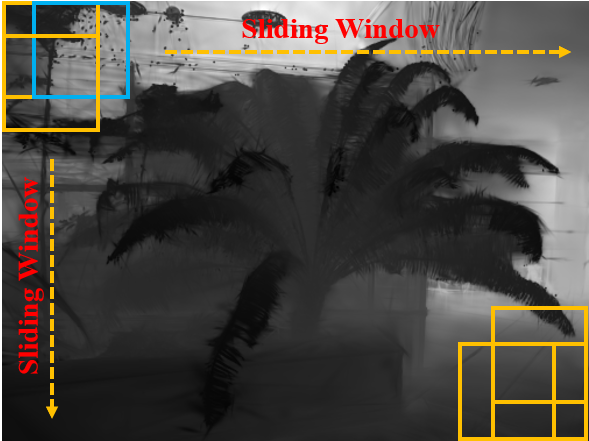}
                \caption{}
            \end{minipage}
            \begin{minipage}[t]{0.15\linewidth}
                \centering
                \includegraphics[width=1\linewidth]{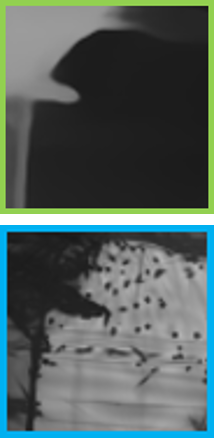}
            \end{minipage}
            \begin{minipage}[t]{0.41\linewidth}
                \centering
                \includegraphics[width=1\linewidth]{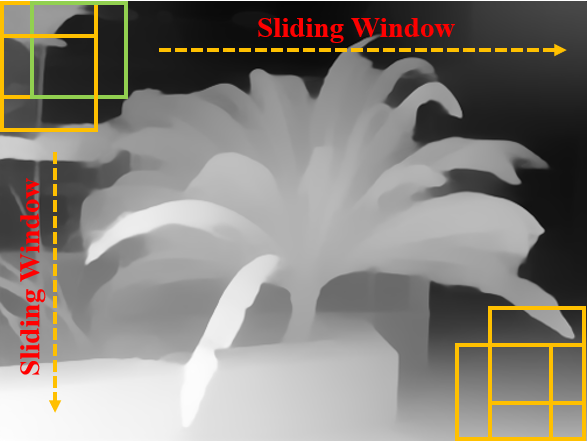}
                \caption{}
            \end{minipage}
        \end{subfigure}

       \caption{Illustration of sliding window-based sampling strategy in DAR. (a) Rendered depth map. (b) Monocular depth map provided by Midas. Our method begins by sliding a window to obtain the rendering depth and mono depth with the specified window size. Instead of computing the \textit{Pearson} loss over the whole image, we compute the region of sliding window areas, which enlarge the \textit{Pearson} loss in the misaligned regions between SfM-derived depth and mono-depth, as illustrated in the blue box of (a) and the green box of (b) of the middle area.}
       \label{fig:depth_slide_window}
\end{figure}

To address this misalignment issue, we develop a sliding window-based depth regularization, as illustrated in Fig. \ref{fig:depth_slide_window}. 
We slide the window from left to right, and from the top to down with over the monocular depth and render depth. We compute the Pearson distribution loss of the window-covered region rather than across the entire image. This region-based Person distribution loss can effectively work with the $L_{1}$ loss generated from sparsely distributed depth values from SfM as they both are effective locally.
Therefore, the final loss function ($L_a$) is formulated as follows: 
\begin{equation}
      L_{a} = \lambda_dL_1(D_p, D_{c}) + \sum_{w=1}^{W}\lambda_p L_p[\chi_w (D_p), \chi_w (D_{m})],
      \label{eq:eq1}
\end{equation}
where $D_p$, $D_c$, and $D_m$ denote the rendered depth, SfM-derived depth, and mono-depth, respectively. $\chi_w$ represents the $w$-th window, and $W$ is the total number of windows.


This local window-based depth-alignment constraint approach not only effectively enforces the absolute scale depth constraint, but also allows the depth within the window to satisfy the relative scale distribution, thus providing depth-aligned depth constrains for more precise rendered depth as shown in (d) of Fig. \ref{fig:depth_align}.





\subsection{Sparse-friendly Sampling (SFS)} \label{method_SFS} 

\begin{figure}[t!]  
	\centering
 
	\begin{subfigure}{\linewidth}
            \rotatebox[origin=c]{90}{\footnotesize{Ellipsoids}\hspace{-2cm}}
            \begin{minipage}[t]{0.46\linewidth}
                \centering
                \includegraphics[width=1\linewidth]{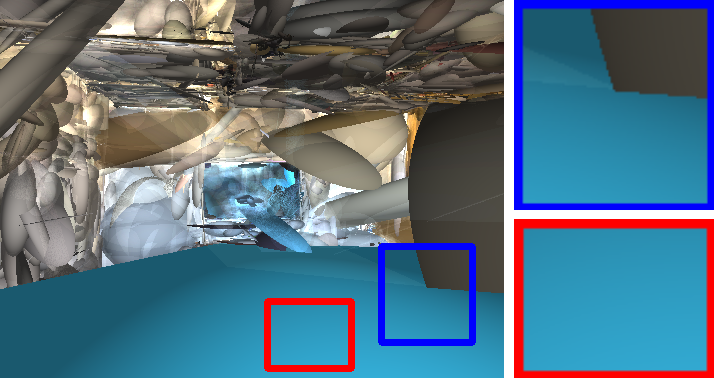}
            \end{minipage}
            \begin{minipage}[t]{0.46\linewidth}
                \centering
                \includegraphics[width=1\linewidth]{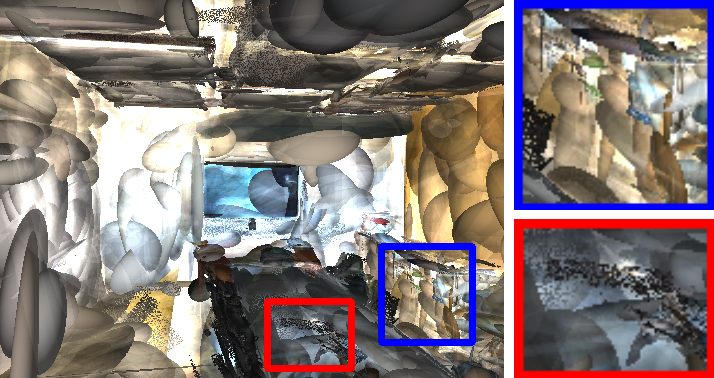}
            \end{minipage}
        \end{subfigure}

        \vspace{0.1cm}
	\begin{subfigure}{\linewidth}
            \rotatebox[origin=c]{90}{\footnotesize{Splats}\hspace{-2cm}}
            \begin{minipage}[t]{0.46\linewidth}
                \centering
                \includegraphics[width=1\linewidth]{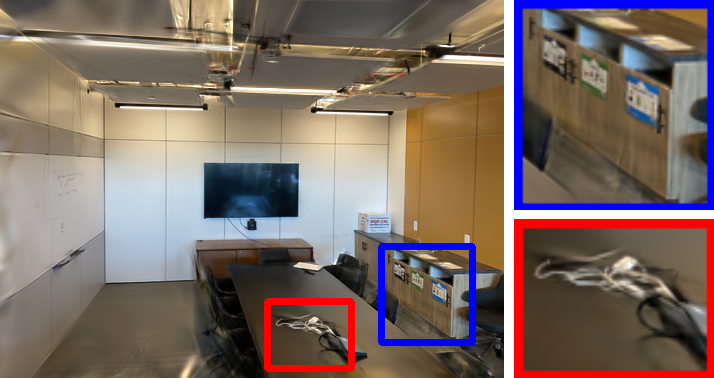}
                 \caption{w/o SFS}
            \end{minipage}
            \begin{minipage}[t]{0.46\linewidth}
                \centering
                \includegraphics[width=1\linewidth]{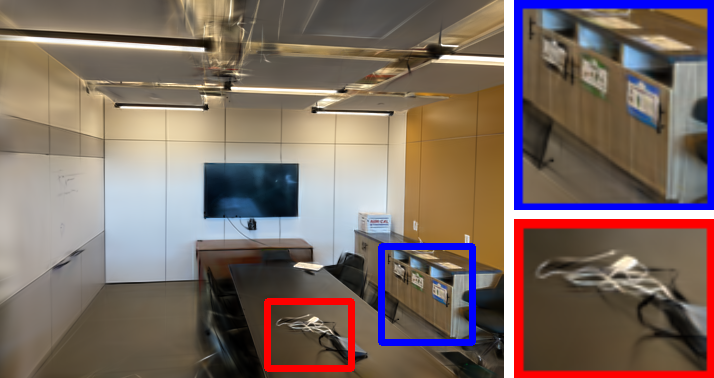}
                \caption{w/ SFS}
            \end{minipage}
        \end{subfigure}

       \caption{Illustration of ellipsoids and their corresponding rendered results. }
       \label{fig:strategy_SFS}
\end{figure}

The Gaussian ellipsoid densification scheme in 3DGS performs effectively with an large number of training images. However, with extremely sparse input data, some Gaussian ellipsoids may grow excessively large with extremely sparse input data and lead to inferior rendered results, as illustrated in (a) of Fig. \ref{fig:strategy_SFS}. This is caused by two primary factors: 
1) The initial scale or size of Gaussian ellipsoids is determined by the average distance from its three nearest neighbours. In cases where the sparse initialized point cloud fully covers the entire space, this method can lead to irrationally large scales for some ellipsoids. 
2) The extreme sparsity of input data can cause rapid increases in some ellipsoids' scale along certain directions. This leads to overfitting the training views while significantly deteriorating the performance of novel viewing perspective. 


One obvious solution is to increase the frequency of Gaussian densification or lower the Gaussian densification threshold. However, these direct strategies would exacerbate the overfitting of Gaussian Splatting in sparse input settings. To address it,
we introduce a non-trivial strategy, named \textbf{Sparse-friendly Sampling}, which selectively applies densification to Gaussian ellipsoids that adversely affect rendering. In this way, we enhance the representation of Gaussian ellipsoids of large PSNR error area by splitting the oversized Gaussian ellipsoids to produce more Gaussian ellipsoids. 

 Figure. \ref{fig:strategy_SFS} (a) shows that the oversized ellipsoids could lead to the pixels blurring, resulting in lower PSNR values. Based on such observations, we traverse all rendered pixels of the training views and collect pixels with the highest PSNR errors. Since the color of each pixel is derived from multiple ellipsoid primitives through $\alpha$-blending, we identify the ellipsoid with the largest weight $\omega$ (calculated by $\omega = \alpha_i \prod_{j=1}^{i-1} (1 - \alpha_j)$) as the primary determinant of the pixel's final color. Subsequently, we apply a splitting procedure to the Gaussian primitives, akin to the method employed in the original 3D Gaussian Splatting (3DGS). This process subdivides large ellipsoids into $m$ smaller ellipsoids to enhance the representation of fine details (in our experiments, $m = 2$).
Besides, we introduce opacity regularization during the training process to encourage non-maximum weight ellipsoids along a camera ray to be more transparent. These operations can reduce the number of excessively large ellipsoids and enhance detailed geometric and rendered results as shown in (b) of Fig. \ref{fig:strategy_SFS}.






\subsection{Optimization} 
We summarize our training constraints as follows:
\begin{equation}
      \mathcal{L} =  \lambda_{1}L_1(C_p,C_{gt})+\lambda_{2}L_{\text{D-SSIM}}(C_p,C_{gt})+L_a+L_o,
\end{equation}
where $C_p$ and $C_{gt}$ denote the rendered and GT images, respectively. $L_1$ and $L_{\text{D-SSIM}}$ represent the photometric and SSIM loss. $L_a$ is the depth-alignment loss computed as Eq. \ref{eq:eq1}. $L_o$ is the non-maximum weight regularization: 
\begin{equation}
      L_o=\frac{\lambda_o}{N} \sum_{n=1}^{N}\left |\alpha_n  \right |,
\end{equation}
where $N$ denotes the total number of non-maximum weighted ellipses hit by all pixels in an image.
In addition, we compute $L_1$, $L_{\text{D-SSIM}}$ and $L_o$ for training views and $L_a$ for both training views and pseudo views.

\section{Experiment}
\label{sec:Results}

\subsection{Experimental Settings}
\textbf{Datasets.} 
To evaluate our sparse-input method, we conduct experiments on four widely used sparse-view datasets for novel view synthesis: LLFF \cite{mildenhall2019local}, DTU \cite{jensen2014large}, Mip-NeRF360 \cite{barron2022mip}, and Blender \cite{mildenhall2020nerf}. 
The LLFF dataset \cite{mildenhall2019local} includes eight forward-facing scenes. 
Following previous methods \cite{zhu2023fsgs,li2024dngaussian}, we select every eighth image as the held-out testing view, and evenly sample sparse views from the remaining images as the training views. 
For each scene, three views are utilized to train all the approaches. 
During evaluation, the image resolution are set to 1008 $\times$ 756 and 504 $\times$ 378. 

The Mip-NeRF360 dataset \cite{barron2022mip} consists of nine scenes, each containing a complex central area or object against an intricate background. 
As per the protocol in \cite{zhu2023fsgs}, we utilize seven of these scenes for our experiments, employing 24 views with images downsampling rates of 4 and 8 for training all methods. 

The DTU dataset \cite{jensen2014large} is a comprehensive object-level dataset. In line with \cite{wang2023sparsenerf,li2024dngaussian}, we use the same 15 scenes with three views for training in the experiments. 
Consistent with the evaluation protocol of prior research \cite{niemeyer2022regnerf,li2024dngaussian}, object masks are employed to exclude the background during inference, as evaluating the entire image introduces bias due to background elements. 

The Blender dataset \cite{mildenhall2020nerf} includes eight objects rendered with photorealistic images using Blender. Following \cite{niemeyer2022regnerf,zhu2023fsgs}, we use 8 images for training and 25 unseen images for testing.


\begin{table*}[!t]
\renewcommand{\arraystretch}{1.1}
\caption{Quantitative comparison on LLFF dataset \cite{mildenhall2019local}.} 
\label{table:sota_llff}
\large
\centering 
\begin{adjustbox} {width=\linewidth}
\begin{tabular}{c | p{1.2cm}<{\centering}p{1.2cm}<{\centering}p{1.2cm}<{\centering} | p{1.2cm}<{\centering}p{1.2cm}<{\centering}p{1.2cm}<{\centering} | p{1.2cm}<{\centering}p{1.2cm}<{\centering}p{1.2cm}<{\centering} | p{1.2cm}<{\centering}p{1.2cm}<{\centering}p{1.2cm}<{\centering}}
\Xhline{3\arrayrulewidth}
\multirow{2}{*}{Method} & \multicolumn{3}{c|}{3 Views (1/8 Resolution)} & \multicolumn{3}{c|}{3 Views (1/4 Resolution)} & \multicolumn{3}{c|}{6 Views} & \multicolumn{3}{c}{9 Views} \\
& PSNR$\uparrow$ & SSIM$\uparrow$ & LPIPS$\downarrow$ & PSNR$\uparrow$ & SSIM$\uparrow$ & LPIPS$\downarrow$ & PSNR$\uparrow$ & SSIM$\uparrow$ & LPIPS$\downarrow$ & PSNR$\uparrow$ & SSIM$\uparrow$ & LPIPS$\downarrow$\\
\hline
Mip-NeRF \cite{barron2021mip}           & 16.11    & 0.401    & 0.460          & 15.22     & 0.351     & 0.540          & 22.91   & 0.756   & 0.213   & 24.88   & 0.826   & 0.170   \\
DietNeRF \cite{jain2021putting}         & 14.94    & 0.370    & 0.496           & 13.86     & 0.305     & 0.578          & 21.75   & 0.717   & 0.248   & 24.28   & 0.801   & 0.183   \\
RegNeRF  \cite{niemeyer2022regnerf}     & 19.08    & 0.587    & 0.336           & 18.66     & 0.535     & 0.411          & 23.10   & 0.760   & 0.206   & 24.86   & 0.820   & 0.161   \\
FreeNeRF \cite{yang2023freenerf}        & 19.63    & 0.612    & 0.308           & 19.13     & 0.562     & 0.384          & 25.13   & 0.779   & 0.195   & 25.13   & 0.827   & 0.160   \\
SparseNeRF \cite{wang2023sparsenerf}    & 19.86    & 0.624    & 0.328           & 19.07     & 0.564     & 0.392          & 24.97   & 0.784   & 0.202   & 24.97   & 0.834   & 0.158   \\
3DGS  \cite{3dgs}                       & 17.83    & 0.582    & 0.321       & 16.94     & 0.488     & 0.402          & 22.87   & 0.732   & 0.204   & 24.65   & 0.813   & 0.159   \\
DNGaussian \cite{li2024dngaussian}     & 19.12    & 0.591    & 0.294           & 18.47      & 0.578      & 0.330                & 22.18   & 0.755   & 0.198   & 23.17   & 0.788   & 0.180   \\
FSGS \cite{zhu2023fsgs}                 & 20.43    & 0.682    & 0.248       & 19.71     & 0.642     & 0.283          & 24.20   & 0.811   & 0.173   & 25.32   & 0.856   & 0.136   \\
Ours                                    
& \textbf{20.85}  & \textbf{0.717}   & \textbf{0.205}    
& \textbf{20.19}  & \textbf{0.680}   & \textbf{0.274}   
& \textbf{24.58}  & \textbf{0.827}   & \textbf{0.125}        
& \textbf{25.86}  & \textbf{0.862}   & \textbf{0.103}   \\ 
\Xhline{3\arrayrulewidth}
\end{tabular}
\end{adjustbox}
\end{table*}

\begin{figure*}[ht!]
	\centering
        
	\begin{subfigure}{\linewidth}
            \rotatebox[origin=c]{90}{\footnotesize{Flower}\hspace{-2.2cm}}
            \begin{minipage}[t]{0.193\linewidth}
                \centering
                \includegraphics[width=1\linewidth]{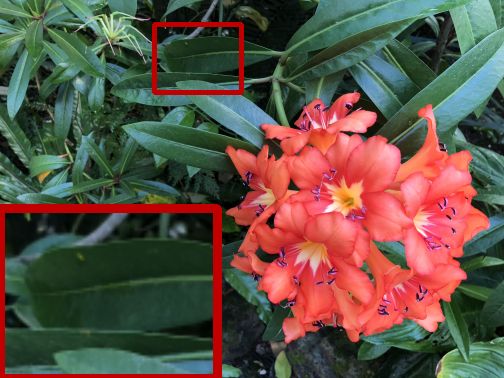}
            \end{minipage}
            \begin{minipage}[t]{0.193\linewidth}
                \centering
                \includegraphics[width=1\linewidth]{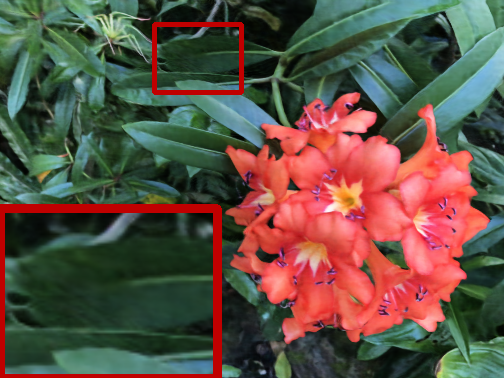}
            \end{minipage}
            \begin{minipage}[t]{0.193\linewidth}
                \centering
                \includegraphics[width=1\linewidth]{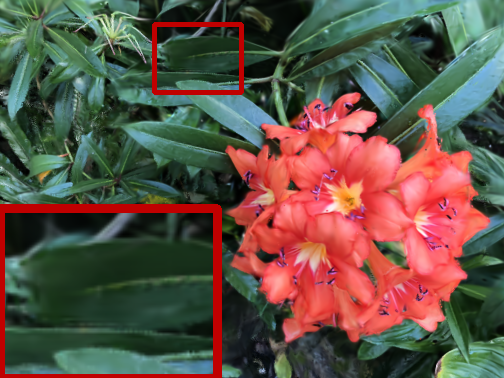}
            \end{minipage}
            \begin{minipage}[t]{0.193\linewidth}
                \centering
                \includegraphics[width=1\linewidth]{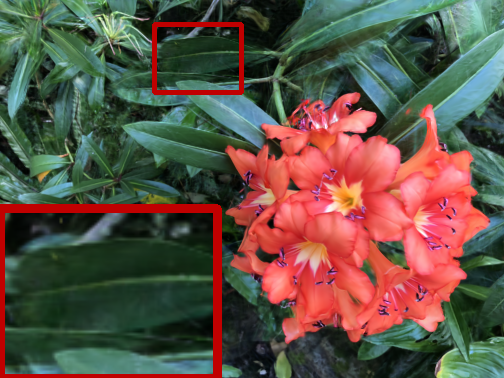}
            \end{minipage}
            \begin{minipage}[t]{0.193\linewidth}
                \centering
                \includegraphics[width=1\linewidth]{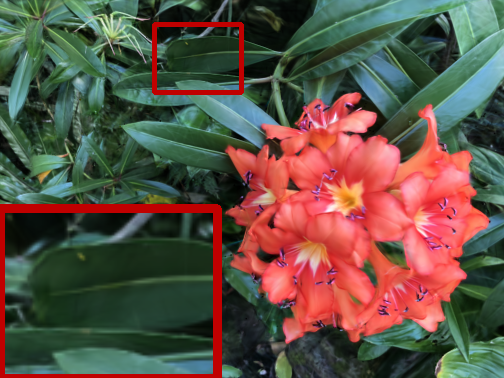}
            \end{minipage}
        \end{subfigure}

	\begin{subfigure}{\linewidth}
            \rotatebox[origin=c]{90}{\footnotesize{Fortress}\hspace{-2.2cm}}
            \begin{minipage}[t]{0.193\linewidth}
                \centering
                \includegraphics[width=1\linewidth]{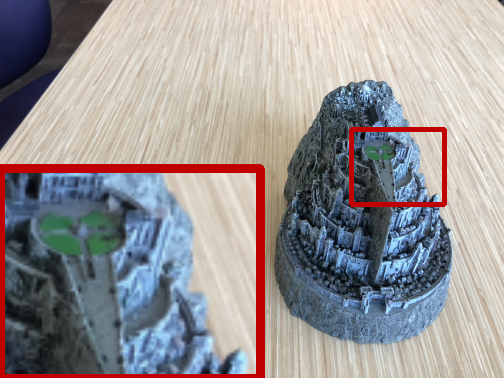}
            \end{minipage}
            \begin{minipage}[t]{0.193\linewidth}
                \centering
                \includegraphics[width=1\linewidth]{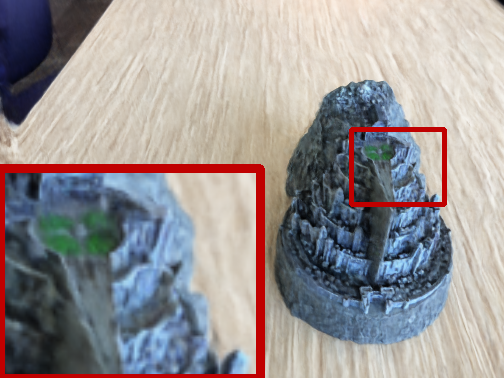}
            \end{minipage}
            \begin{minipage}[t]{0.193\linewidth}
                \centering
                \includegraphics[width=1\linewidth]{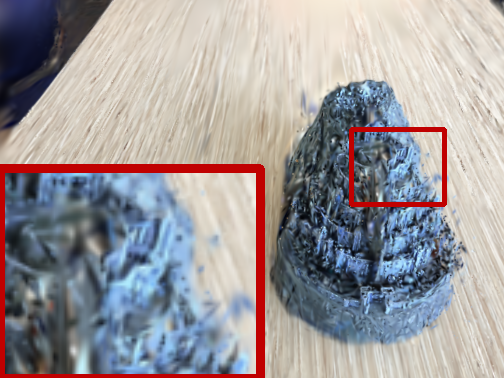}
            \end{minipage}
            \begin{minipage}[t]{0.193\linewidth}
                \centering
                \includegraphics[width=1\linewidth]{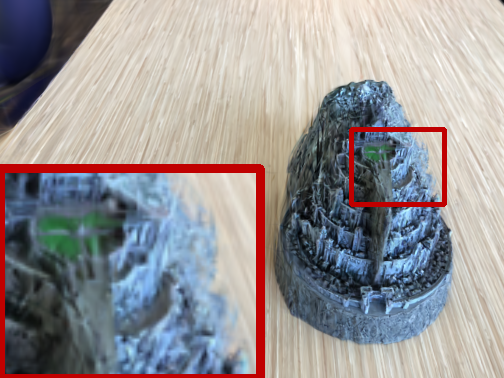}
            \end{minipage}
            \begin{minipage}[t]{0.193\linewidth}
                \centering
                \includegraphics[width=1\linewidth]{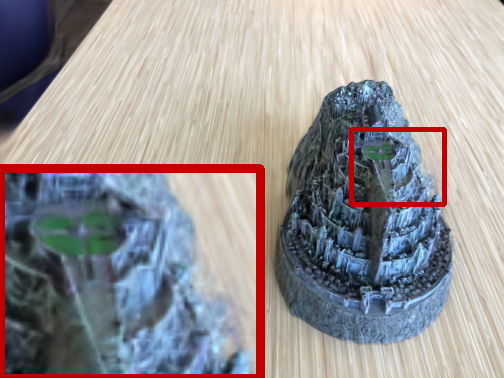}
            \end{minipage}
        \end{subfigure}

	\begin{subfigure}{\linewidth}
            \rotatebox[origin=c]{90}{\footnotesize{Room}\hspace{-2.2cm}}
            \begin{minipage}[t]{0.193\linewidth}
                \centering
                \includegraphics[width=1\linewidth]{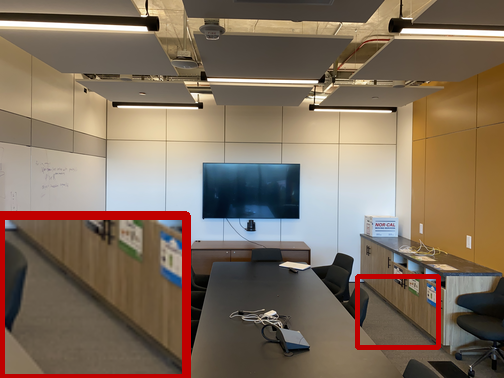}
                \caption{GT}
            \end{minipage}
            \begin{minipage}[t]{0.193\linewidth}
                \centering
                \includegraphics[width=1\linewidth]{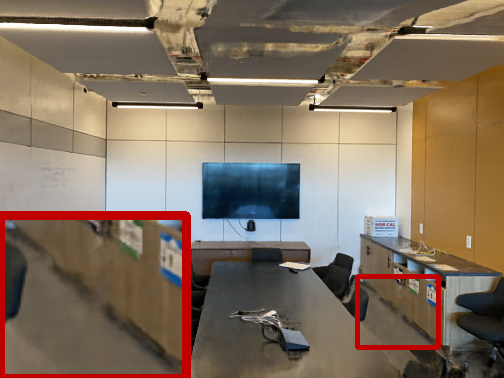}
                \caption{FreeNeRF}
            \end{minipage}
            \begin{minipage}[t]{0.193\linewidth}
                \centering
                \includegraphics[width=1\linewidth]{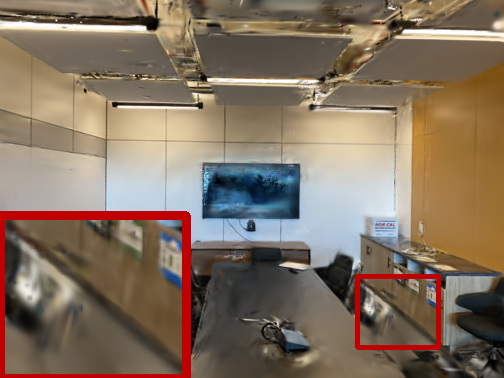}
                \caption{DNGaussian}
            \end{minipage}
            \begin{minipage}[t]{0.193\linewidth}
                \centering
                \includegraphics[width=1\linewidth]{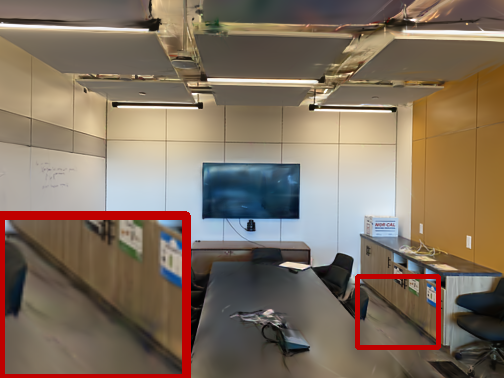}
                \caption{FSGS}
            \end{minipage}
            \begin{minipage}[t]{0.193\linewidth}
                \centering
                \includegraphics[width=1\linewidth]{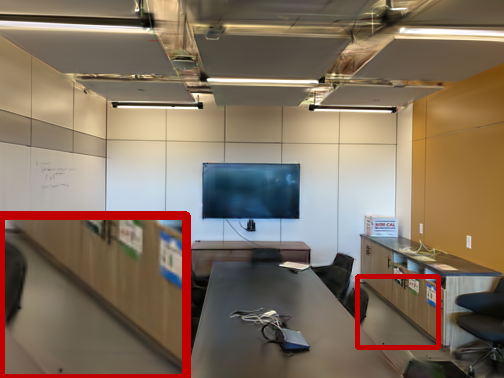}
                \caption{Ours}
            \end{minipage}
        \end{subfigure}

	\caption{Qualitative Results on LLFF Datasets. Our method can produce photorealistic results with finer details. 
    }
	\label{fig:LLFF}
\end{figure*}

\textbf{Evaluation Metrics.} 
For quantitative comparisons, we adopt three evaluation metrics: peak signal-to-noise ratio (PSNR), structural similarity index (SSIM) \cite{wang2004image}, learned perceptual image patch similarity (LPIPS) to assess the visual quality\cite{zhang2018unreasonable}. 

\textbf{Implementation Details.}
Our approach is implemented using the PyTorch framework, and utilizes the pre-trained Midas Model \cite{MiDaS,ranftl2021vision} for zero-shot monocular depth estimation and the Colmap model for initial camera poses and 3D points. We set the total number of loop iterations as 3 and the number of pseudo cameras generated around one training camera $l$ as 4. In a single loop iteration, we start the densification of Gaussian ellipsoids after 1000 iterations, and the frequency of densification is set to 200 iterations. The 2D grad threshold of densification is set to 0.0005.
We set the window length to 32 and the step size of the window sliding to 4 for all datasets.
For each loop, 3DGS are trained with 10,000 iterations. The weights of the loss function $\lambda_1$, $\lambda_2$, $\lambda_o$, $\lambda_d$, $\lambda_p$ are set to 0.8, 0.2, 0.05, 0.005 and 0.05, respectively.

\textbf{Efficiency.} 
Our method achieves average rendering speeds (ARS) of 418, 490, 831, and 720 FPS for the LLFF (3 training views), Mip-NeRF360 (24 training views), DTU (3 training views), and Blender (8 training views) datasets, respectively. 
Here, the average speed is calculated by  $ARS=\frac{1}{20}\sum_{i=1}^{20}  \frac{M}{T} $, where $T$ is the total inference time of all test images across all scenes in one dataset, the $M$ is the number of all test images, and we tested $20$ times to eliminate random error.
The training time of the model depends on the number of loops, with a training time of about 10 min per loop and a total training time of about 35-45 min on all datasets.
All experiment times were evaluated on a 3090 NVIDIA GPU.

\subsection{Comparison With Existing Methods}
\textbf{Comparisons on LLFF.}
As shown in Table \ref{table:sota_llff}, our proposed method outperforms other state-of-the-art approaches across different image resolutions and sparse-input settings. 
Specifically, our approach surpasses the second-best FSGS by 0.42 and 0.38 in PSNR at two test resolutions when only using 3 sparse-input views for training. 
Moreover, we can see that more training views can bring better reconstruction quality and our LoopSparseGS delivers superior performance compared to all other methods, validating the effectiveness of our proposed strategies. 
We show the qualitative results in Fig. \ref{fig:LLFF}. 
Existing methods tend to produce artifact and blurry rendered results. In comparison, our approach exhibits fine-grained details such as the leaves (Scens: \textit{Flower}) and the floor (Scens: \textit{Room}). 

\begin{table*}[!t]
\renewcommand{\arraystretch}{1.1}
\caption{Quantitative comparison on Mip-NeRF360 dataset \cite{barron2022mip}.} 
\label{table:sota_mipnerf}
\centering 
\begin{adjustbox} {width=.7\linewidth}
\begin{tabular}{ p{3.cm}<{\centering} | p{1.25cm}<{\centering}p{1.25cm}<{\centering}p{1.25cm}<{\centering} | p{1.25cm}<{\centering}p{1.25cm}<{\centering}p{1.25cm}<{\centering}  }
\Xhline{3\arrayrulewidth}
\multirow{2}{*}{Method} & \multicolumn{3}{c|}{24 Views (1/8 Resolution)} & \multicolumn{3}{c}{24 Views (1/4 Resolution)}  \\
& PSNR$\uparrow$ & SSIM$\uparrow$ & LPIPS$\downarrow$ & PSNR$\uparrow$ & SSIM$\uparrow$ & LPIPS$\downarrow$  \\
\hline
Mip-NeRF \cite{barron2021mip}           & 21.23      & 0.613     & 0.351        & 19.78      & 0.530     & 0.431     \\
DietNeRF \cite{jain2021putting}         & 20.21      & 0.557     & 0.387        & 19.11      & 0.482     & 0.452     \\
RegNeRF  \cite{niemeyer2022regnerf}     & 22.19      & 0.643     & 0.335        & 20.55      & 0.546     & 0.398     \\
FreeNeRF \cite{yang2023freenerf}        & 22.78      & 0.689     & 0.323        & 21.39      & 0.587     & 0.377     \\
SparseNeRF \cite{wang2023sparsenerf}    & 22.85      & 0.693     & 0.315        & 21.43      & 0.604     & 0.389     \\
3DGS  \cite{3dgs}                       & 20.89      & 0.633     & 0.317        & 19.93      & 0.588     & 0.401     \\
DNGaussian \cite{li2024dngaussian}      & 22.00      & 0.683     & 0.287        & 21.93      &    0.668     & 0.337     \\
FSGS \cite{zhu2023fsgs}                 & 23.70      & 0.745     & 0.230        & 22.52      & 0.673     & 0.313     \\
Ours                                    
& \textbf{24.09} & \textbf{0.755} & \textbf{0.226} 
& \textbf{23.54} & \textbf{0.722} & \textbf{0.288} \\ 
\Xhline{3\arrayrulewidth}
\end{tabular}
\end{adjustbox}
\end{table*}

\begin{figure*}[ht!]
	\centering
        

	\begin{subfigure}{\linewidth}
            \rotatebox[origin=c]{90}{\footnotesize{Bonsai}\hspace{-2.2cm}}
            \begin{minipage}[t]{0.193\linewidth}
                \centering
                \includegraphics[width=1\linewidth]{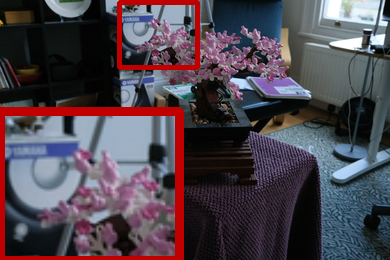}
            \end{minipage}
            \begin{minipage}[t]{0.193\linewidth}
                \centering
                \includegraphics[width=1\linewidth]{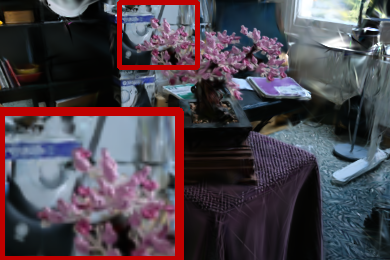}
            \end{minipage}
            \begin{minipage}[t]{0.193\linewidth}
                \centering
                \includegraphics[width=1\linewidth]{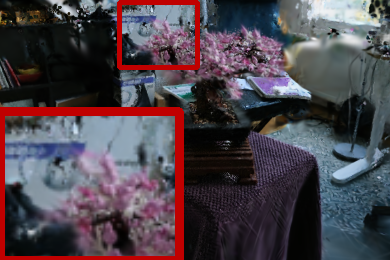}
            \end{minipage}
            \begin{minipage}[t]{0.193\linewidth}
                \centering
                \includegraphics[width=1\linewidth]{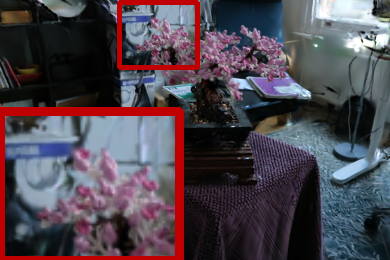}
            \end{minipage}
            \begin{minipage}[t]{0.193\linewidth}
                \centering
                \includegraphics[width=1\linewidth]{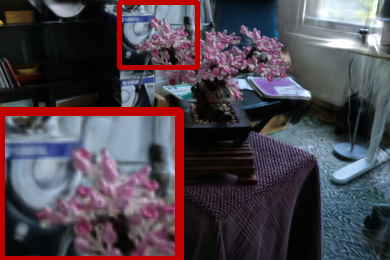}
            \end{minipage}
        \end{subfigure}

	\begin{subfigure}{\linewidth}
            \rotatebox[origin=c]{90}{\footnotesize{Counter}\hspace{-2.2cm}}
            \begin{minipage}[t]{0.193\linewidth}
                \centering
                \includegraphics[width=1\linewidth]{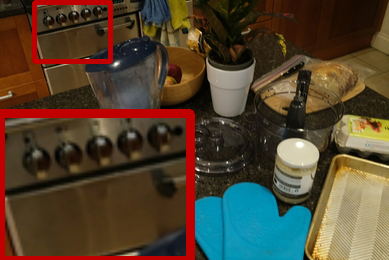}
            \end{minipage}
            \begin{minipage}[t]{0.193\linewidth}
                \centering
                \includegraphics[width=1\linewidth]{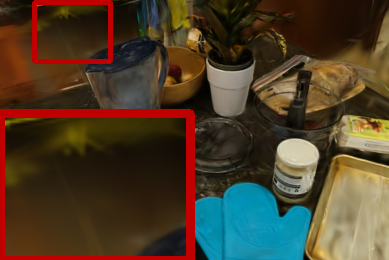}
            \end{minipage}
            \begin{minipage}[t]{0.193\linewidth}
                \centering
                \includegraphics[width=1\linewidth]{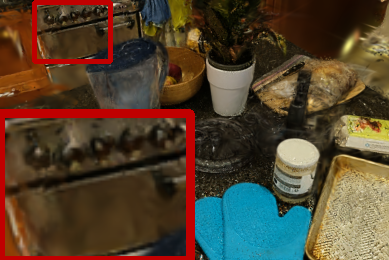}
            \end{minipage}
            \begin{minipage}[t]{0.193\linewidth}
                \centering
                \includegraphics[width=1\linewidth]{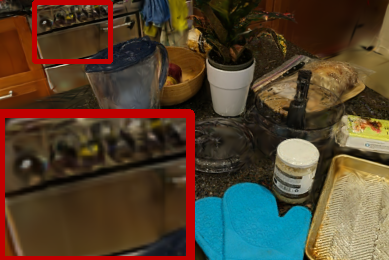}
            \end{minipage}
            \begin{minipage}[t]{0.193\linewidth}
                \centering
                \includegraphics[width=1\linewidth]{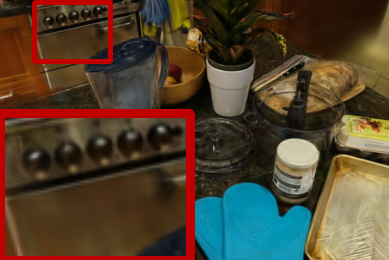}
            \end{minipage}
        \end{subfigure}

	\begin{subfigure}{\linewidth}
            \rotatebox[origin=c]{90}{\footnotesize{Garden}\hspace{-2.2cm}}
            \begin{minipage}[t]{0.193\linewidth}
                \centering
                \includegraphics[width=1\linewidth]{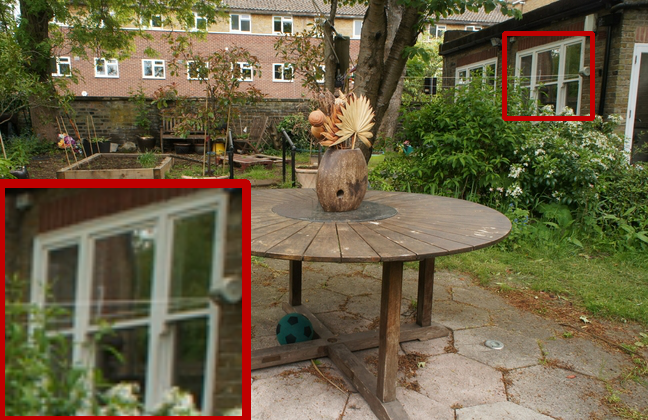}
                \caption{GT}
            \end{minipage}
            \begin{minipage}[t]{0.193\linewidth}
                \centering
                \includegraphics[width=1\linewidth]{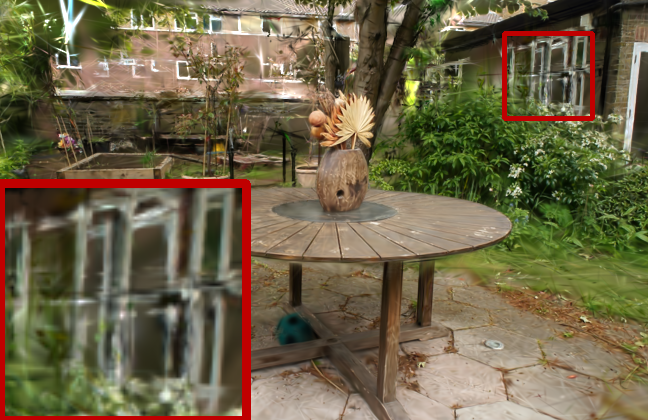}
                \caption{3DGS}
            \end{minipage}
            \begin{minipage}[t]{0.193\linewidth}
                \centering
                \includegraphics[width=1\linewidth]{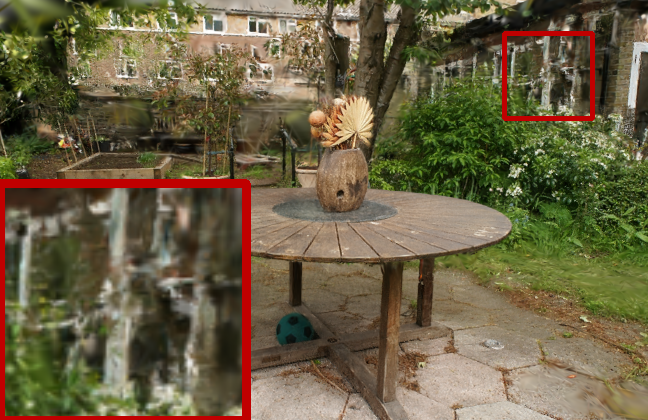}
                \caption{DNGaussian}
            \end{minipage}
            \begin{minipage}[t]{0.193\linewidth}
                \centering
                \includegraphics[width=1\linewidth]{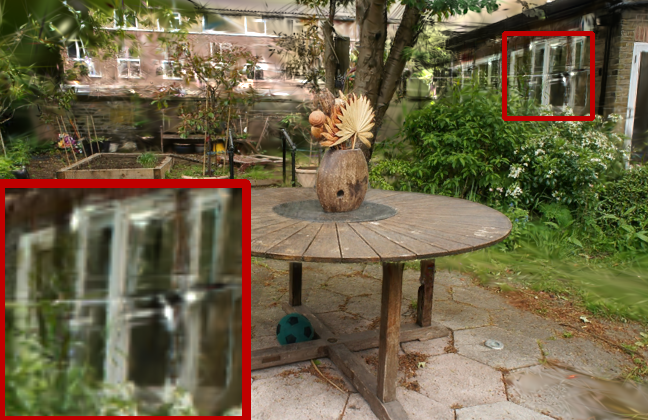}
                \caption{FSGS}
            \end{minipage}
            \begin{minipage}[t]{0.193\linewidth}
                \centering
                \includegraphics[width=1\linewidth]{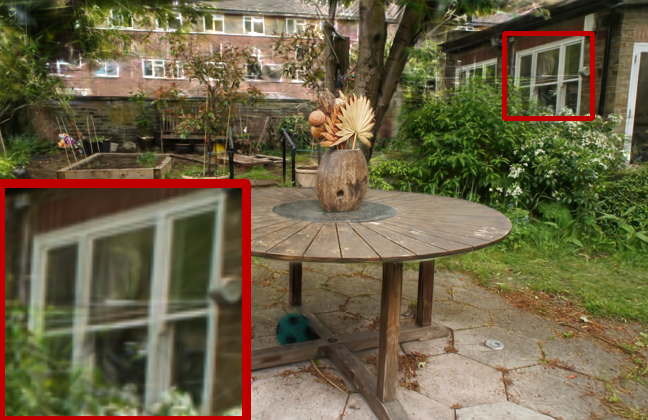}
                \caption{Ours}
            \end{minipage}
        \end{subfigure}

	\caption{Qualitative Results on Mip-NeRF360 Datasets. Our approach can render photorealistic results with more complete structures and finer details. 
            }
	\label{fig:mip360}
\end{figure*}

\textbf{Comparisons on Mip-NeRF360.}
Table \ref{table:sota_mipnerf} presents the quantitative results in complex scenes from Mip-NeRF360. 
It can be seen that our method also outperforms other state-of-the-art approaches in terms of various metrics across different image resolutions. 
Compared to Mip-NeRF requiring dense-input, although methods using regularizations or depth information for sparse-input, such as FreeNeRF and SparseNeRF, enhance rendering quality to some extent, they still encounter a performance bottleneck. 
Compared to FSGS that incorporates Gaussian unpooling densification technique and monocular depth maps, our proposed method significantly outperforms it with an improvement of 0.39 and 1.02 in PSNR across two resolutions. These demonstrate the effectiveness of our proposed loop-based mechanism and importance-guided sampling strategy.   
Moreover, we provide qualitative comparison in Fig. \ref{fig:mip360}. It can be seen that existing methods tend to produce blurred rendered results with incomplete structure, as highlighted by the red boxes around the ``plate," ``switch," and ``window." 
In contrast, our proposed loop-based and sparse-friendly sampling strategies yield denser initialized Gaussians with effective geometric constraints, resulting in more complete structures and finer details.

\begin{figure}[t!]
	\centering
 
	\begin{subfigure}{\linewidth}
            \rotatebox[origin=c]{90}{\footnotesize{GT}\hspace{-1.4cm}}
            \begin{minipage}[t]{0.31\linewidth}
                \centering
                \includegraphics[width=1\linewidth]{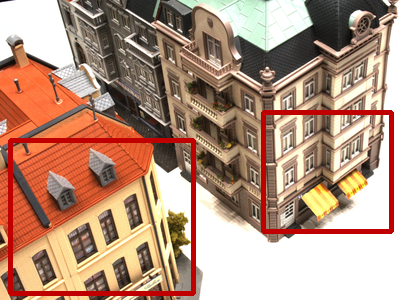}
            \end{minipage}
            \begin{minipage}[t]{0.31\linewidth}
                \centering
                \includegraphics[width=1\linewidth]{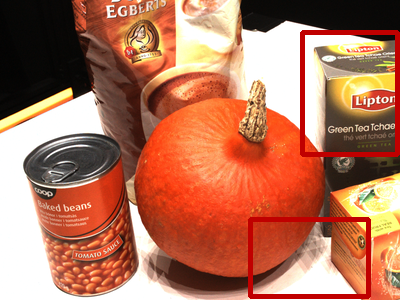}
            \end{minipage}
            \begin{minipage}[t]{0.31\linewidth}
                \centering
                \includegraphics[width=1\linewidth]{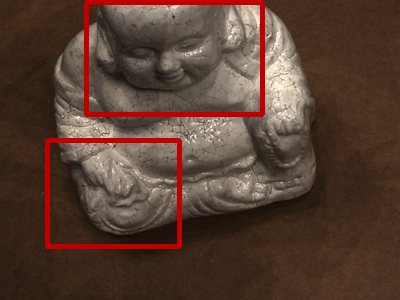}
            \end{minipage}
        \end{subfigure}

	\begin{subfigure}{\linewidth}
            \rotatebox[origin=c]{90}{\footnotesize{FreeNeRF}\hspace{-1.4cm}}
            \begin{minipage}[t]{0.31\linewidth}
                \centering
                \includegraphics[width=1\linewidth]{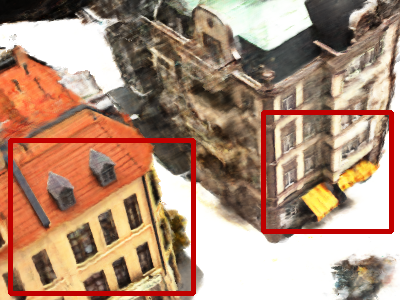}
            \end{minipage}
            \begin{minipage}[t]{0.31\linewidth}
                \centering
                \includegraphics[width=1\linewidth]{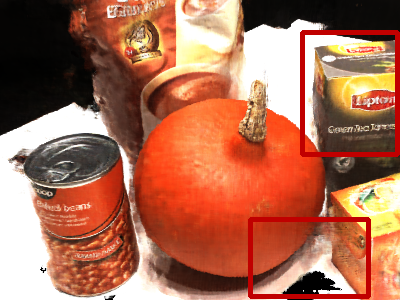}
            \end{minipage}
            \begin{minipage}[t]{0.31\linewidth}
                \centering
                \includegraphics[width=1\linewidth]{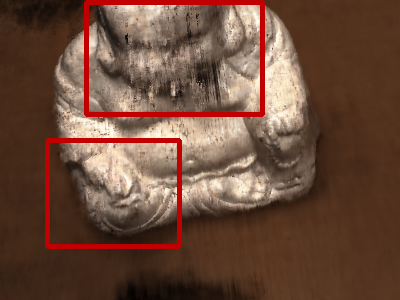}
            \end{minipage}
        \end{subfigure}

	\begin{subfigure}{\linewidth}
            \rotatebox[origin=c]{90}{\footnotesize{DNGaussian}\hspace{-1.4cm}}
            \begin{minipage}[t]{0.31\linewidth}
                \centering
                \includegraphics[width=1\linewidth]{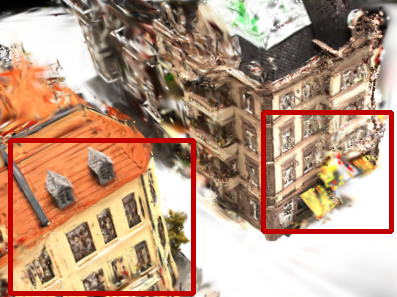}
            \end{minipage}
            \begin{minipage}[t]{0.31\linewidth}
                \centering
                \includegraphics[width=1\linewidth]{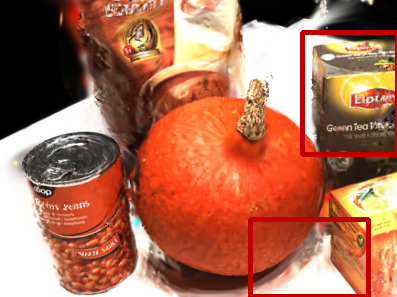}
            \end{minipage}
            \begin{minipage}[t]{0.31\linewidth}
                \centering
                \includegraphics[width=1\linewidth]{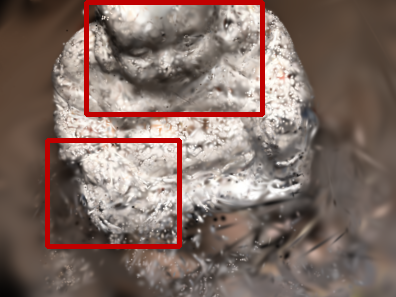}
            \end{minipage}
        \end{subfigure}

	\begin{subfigure}{\linewidth}
            \rotatebox[origin=c]{90}{\footnotesize{Ours}\hspace{-1.4cm}}
            \begin{minipage}[t]{0.31\linewidth}
                \centering
                \includegraphics[width=1\linewidth]{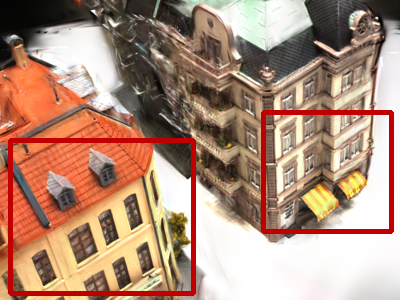}
                \caption{Scan21}
            \end{minipage}
            \begin{minipage}[t]{0.31\linewidth}
                \centering
                \includegraphics[width=1\linewidth]{figure/DTU/scan31_DNGaussian.png}
                \caption{Scan31}
            \end{minipage}
            \begin{minipage}[t]{0.31\linewidth}
                \centering
                \includegraphics[width=1\linewidth]{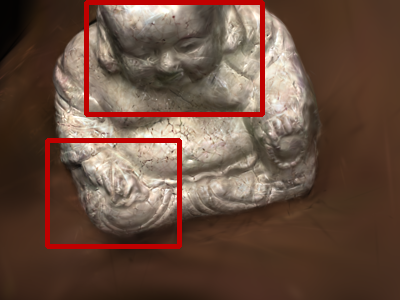}
                \caption{Scan114}
            \end{minipage}
        \end{subfigure}

	\caption{Qualitative results on DTU datasets.}
	\label{fig:DTU}
\end{figure}

\textbf{Comparisons on DTU.}
As shown in Table \ref{table:sota_dtu}, we present the quantitative results on the DTU 3-view sparse-input setting. For the object-level scenes, our method also achieve the best rendering quality in terms of different metrics, with the significant improvement of 0.76 in PSNR. 
In Fig. \ref{fig:DTU}, we present the visual comparisons. 
Compared to other methods that produce blurry renderings, our approach can capture color details much closer to the ground truth, demonstrating its effectiveness on object-level real-world scenes.

\begin{figure}[t!]
	\centering

	\begin{subfigure}{\linewidth}
            \rotatebox[origin=c]{90}{\footnotesize{Ficus}\hspace{-1.4cm}}
            \begin{minipage}[t]{0.31\linewidth}
                \centering
                \includegraphics[width=1\linewidth]{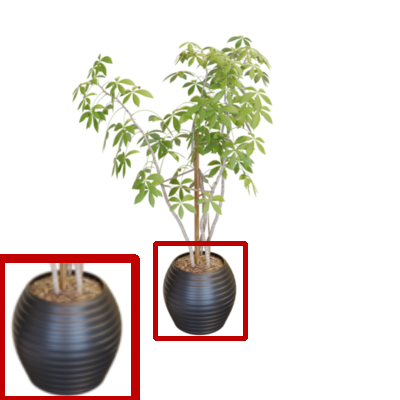}
            \end{minipage}
            \begin{minipage}[t]{0.31\linewidth}
                \centering
                \includegraphics[width=1\linewidth]{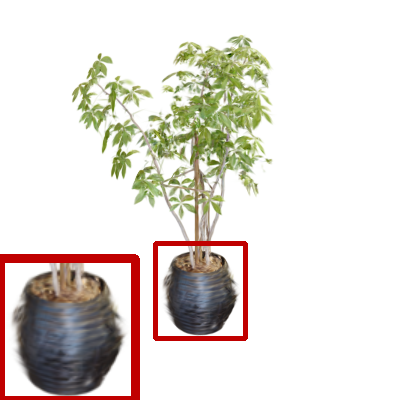}
            \end{minipage}
            \begin{minipage}[t]{0.31\linewidth}
                \centering
                \includegraphics[width=1\linewidth]{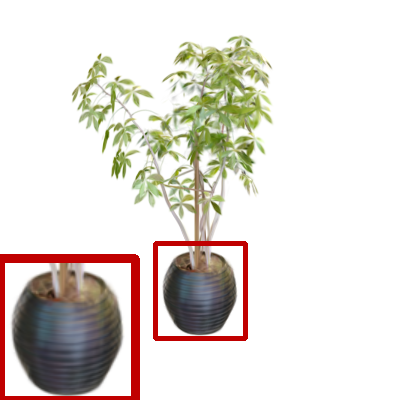}
            \end{minipage}
        \end{subfigure}

	\begin{subfigure}{\linewidth}
            \rotatebox[origin=c]{90}{\footnotesize{Drums}\hspace{-1.4cm}}
            \begin{minipage}[t]{0.31\linewidth}
                \centering
                \includegraphics[width=1\linewidth]{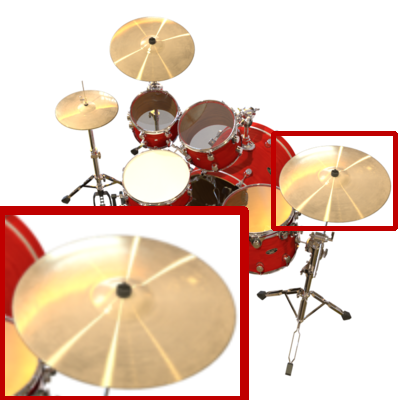}
                \caption{GT}
            \end{minipage}
            \begin{minipage}[t]{0.31\linewidth}
                \centering
                \includegraphics[width=1\linewidth]{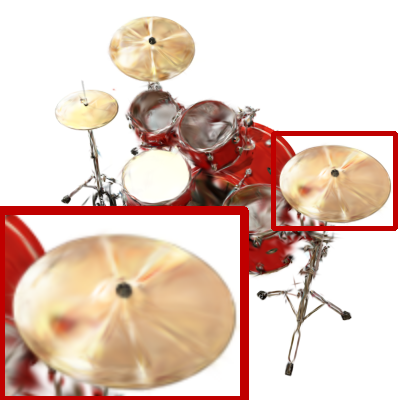}
                \caption{DNGaussian}
            \end{minipage}
            \begin{minipage}[t]{0.31\linewidth}
                \centering
                \includegraphics[width=1\linewidth]{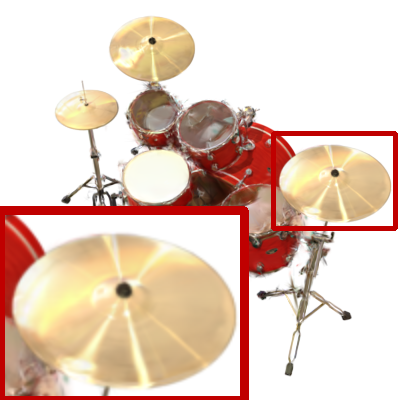}
                \caption{Ours}
            \end{minipage}
        \end{subfigure}


	\caption{Qualitative results on Blender datasets.}
	\label{fig:Blender}
\end{figure}

\textbf{Comparisons on Blender.}
Table \ref{table:sota_blender} shows the quantitative results on the Blender dataset with an 8-view sparse-input setting. 
Our method also significantly outperforms other approaches on the synthesis scenes, achieving a 0.92 improvement in PSNR compared to the second-best method. 
Furthermore, Fig. \ref{fig:Blender} illustrates the rendered results. 
DNGaussian struggles to achieve precise texture and illumination, whereas our approach accurately captures the geometry of objects and authentic reflections. This demonstrates the effectiveness of our proposed strategies for object-level synthesis scenes.

\begin{table}[!t]
\renewcommand{\arraystretch}{1.1}
\caption{Quantitative comparison on DTU dataset \cite{jensen2014large}.} 
\label{table:sota_dtu}
\centering 
\begin{adjustbox} {width=.8\linewidth}
\begin{tabular}{c | p{1.25cm}<{\centering}p{1.25cm}<{\centering}p{1.25cm}<{\centering} }
\Xhline{3\arrayrulewidth}
\multirow{2}{*}{Method} & \multicolumn{3}{c}{3 Views}\\
& PSNR$\uparrow$ & SSIM$\uparrow$ & LPIPS$\downarrow$ \\ 
\hline
Mip-NeRF \cite{barron2021mip}           & 8.68  & 0.571          & 0.353           \\
DietNeRF \cite{jain2021putting}         & 11.85 & 0.633          & 0.314           \\
RegNeRF  \cite{niemeyer2022regnerf}     & 18.89 & 0.745          & 0.190           \\
FreeNeRF \cite{yang2023freenerf}        & 19.92 & 0.787          & 0.182           \\
SparseNeRF \cite{wang2023sparsenerf}    & 19.55 & 0.769          & 0.201           \\
3DGS  \cite{3dgs}                       & 10.99 & 0.585          & 0.313           \\
DNGaussian \cite{li2024dngaussian}     & 18.91 & 0.790          & 0.176           \\
Ours                                    & \textbf{20.68} & \textbf{0.856} & \textbf{0.125} \\
\Xhline{3\arrayrulewidth}
\end{tabular}
\end{adjustbox}
\end{table}

\begin{table}[!t]
\renewcommand{\arraystretch}{1.1}
\caption{Quantitative comparison on Blender dataset \cite{mildenhall2020nerf}.} 
\label{table:sota_blender}
\centering 
\begin{adjustbox} {width=.8\linewidth}
\begin{tabular}{c | p{1.25cm}<{\centering}p{1.25cm}<{\centering}p{1.25cm}<{\centering} }
\Xhline{3\arrayrulewidth}
\multirow{2}{*}{Method} & \multicolumn{3}{c}{8 Views}\\
& PSNR$\uparrow$ & SSIM$\uparrow$ & LPIPS$\downarrow$  \\ 
\hline
Mip-NeRF \cite{barron2021mip}           & 20.89   & 0.830   & 0.168    \\
DietNeRF \cite{jain2021putting}         & 22.50   & 0.823   & 0.124    \\
RegNeRF  \cite{niemeyer2022regnerf}     & 23.86   & 0.852   & 0.105     \\
FreeNeRF \cite{yang2023freenerf}        & 24.26   & 0.883   & 0.098     \\
SparseNeRF \cite{wang2023sparsenerf}    & 24.04   & 0.876   & 0.113     \\
3DGS  \cite{3dgs}                       & 21.56   & 0.847   & 0.130     \\
DNGaussian \cite{li2024dngaussian}     & 24.31   & 0.886   & 0.088     \\
FSGS  \cite{zhu2023fsgs}                & 24.64   & 0.895   & 0.095     \\
Ours                                    & \textbf{25.56}   & \textbf{0.906}   & \textbf{0.075} \\
\Xhline{3\arrayrulewidth}
\end{tabular}
\end{adjustbox}
\end{table}

\begin{table}[!t]
\renewcommand{\arraystretch}{1.2}
\caption{Ablation study for our LoopSparceGS.}
\label{table:Ablation_all}
\centering
\begin{adjustbox} {width=.98\linewidth}
\begin{tabular}{ c | c c c | p{1.25cm}<{\centering} p{1.25cm}<{\centering} p{1.25cm}<{\centering}}
\Xhline{3\arrayrulewidth}
Index     & DAR & PGI & SFS    & PSNR $\uparrow$  & SSIM $\uparrow$ & LPIPS $\downarrow$ \\
\hline
(a)       &            &             &             &  19.137 &  0.637  &  0.247 \\
(b)       & \checkmark &             &             &  19.466 &  0.658  &  0.236 \\
(c)       &            & \checkmark  &             &  19.940 &  0.682  &  0.217 \\
(d)       &            &             & \checkmark  &  19.419 &  0.638  &  0.250 \\
(e)       & \checkmark & \checkmark  &             &  20.565 &  0.710  &  0.208 \\
(f)       & \checkmark &             & \checkmark  &  20.203 &  0.677  &  0.225 \\
(g)       &            & \checkmark  & \checkmark  &  20.158 &  0.692  &  0.214 \\
(h)       & \checkmark & \checkmark  & \checkmark  &  \textbf{20.846} &  \textbf{0.717}  & \textbf{ 0.205} \\
\Xhline{3\arrayrulewidth}
\end{tabular}
\end{adjustbox} 
\end{table}

\begin{table}[!t]
\renewcommand{\arraystretch}{1.1}
\caption{Ablation study for the Depth-alignment Regularization.} 
\label{table:ablation_depth}
\centering 
\begin{adjustbox} {width=\linewidth}
\begin{tabular}{ c | p{1.25cm}<{\centering} p{1.25cm}<{\centering}p{1.25cm}<{\centering} p{1.25cm}<{\centering} }
\Xhline{3\arrayrulewidth}
{Setting} & PSNR$\uparrow$ & SSIM$\uparrow$ & LPIPS$\downarrow$ \\ 
\hline
wo Depth Loss                  & 20.158   & 0.692   & 0.214  \\ 
SfM Depth Loss              & 20.205   & 0.690   & 0.210  \\
Monocular Depth Loss           & 20.620   & 0.706   & 0.215  \\
SfM + Monocular Depth Loss  & 20.720   & 0.709   & 0.207  \\
Proposed Depth-alignment Loss  & \textbf{20.846} &  \textbf{0.717}  & \textbf{0.205} \\
\Xhline{3\arrayrulewidth}
\end{tabular}
\end{adjustbox}
\end{table}

\subsection{Ablation Experiments}
In this section, we ablate our method on the LLFF 3-view with 503 $\times$ 381 image resolution setting. The quantitative results are presented in Tables \ref{table:Ablation_all} to 
\ref{table:ablation_pseudo_views}.

\subsubsection{Effectiveness of architecture modules}
In Table \ref{table:Ablation_all}, we present the ablation results obtained by progressively applying our Depth-alignment Regularization (DAR), Progressive Gaussian Initialization (PGI), and Sparse-friendly Sampling (SFS) strategies. 
Compared to the baseline (a), each proposed module contributes to improved reconstruction quality, as shown in (b)-(d). Specifically, 
DAR provides additional depth information to mitigate the under-constrained effect caused by sparse-input, thereby enhancing reconstruction quality. 
PGI introduces a loop-based initialization strategy and offers denser Gaussians for scene modeling, resulting in a 0.803 increase in PSNR. 
SFS provides a Gaussian densification strategy suitable for sparse-view training and exhibits a gain of 0.282 PSNR. 
Furthermore, as shown in (e)-(g), the combination of these strategies yields better results, with all three working together to produce the best performance, as demonstrated in (h). 

\subsubsection{Effectiveness of depth-alignment loss}
Table \ref{table:ablation_depth} presents the ablation of our depth-alignment loss through various depth loss configurations. 
The experiments indicate that both SfM depth loss and monocular depth loss enhance the sparse-input reconstruction quality, with their combination yielding even better results. 
Moreover, applying our depth-alignment regularization with window-level Pearson correlation loss can further boost the performance. 
This improvement is attributed to that our window-level operation can effectively align the absolute depth and relative depth constraints and rectify incorrect depth constraints. 

\begin{table}[!t]
\renewcommand{\arraystretch}{1.2}
\caption{Ablation study for different filter strategies in PGI.}
\label{table:Ablation_filter}
\centering
\small
\begin{adjustbox} {width=.95\linewidth}
\begin{tabular}{ c | c c c | p{1.2cm}<{\centering} p{1.2cm}<{\centering} p{1.2cm}<{\centering}}
\Xhline{3\arrayrulewidth}
Index     & Filter1 & Filter2 & Filter3    & PSNR $\uparrow$  & SSIM $\uparrow$ & LPIPS $\downarrow$ \\
\hline
(a)       &            &             &             &  20.240 &  0.697  &  0.222 \\
(b)       & \checkmark &             &             &  20.404 &  0.700  &  0.221 \\
(c)       &            & \checkmark  &             &  20.661 &  0.712  &  0.205 \\
(d)       &            &             & \checkmark  &  20.386 &  0.701  &  0.217 \\
(e)       & \checkmark & \checkmark  &             &  20.744 &  0.715  &  0.205 \\
(f)       & \checkmark &             & \checkmark  &  20.629 &  0.706  &  0.216 \\
(g)       &            & \checkmark  & \checkmark  &  20.700 &  0.716  &  0.202 \\
(h)       & \checkmark & \checkmark  & \checkmark  &  \textbf{20.846} &  \textbf{0.717}  & \textbf{ 0.205} \\
\Xhline{3\arrayrulewidth}
\end{tabular}
\end{adjustbox} 
\end{table}

\begin{table}[!t]
\renewcommand{\arraystretch}{1.1}
\caption{Ablation study for Sparse-friendly Sampling.} 
\label{table:ablation_sampling}
\centering 
\begin{adjustbox} {width=.95\linewidth}
\begin{tabular}{c | p{1.25cm}<{\centering}p{1.25cm}<{\centering}p{1.25cm}<{\centering} p{1.25cm}<{\centering} }
\Xhline{3\arrayrulewidth}
Setting  & PSNR$\uparrow$ & SSIM$\uparrow$ & LPIPS$\downarrow$ \\ 
\hline
w/o Non-maximum regularization & 20.674   & 0.716   & 0.192  \\
w/o Maximum split              & 20.814   & 0.713   & 0.217  \\
Proposed Sparse-friendly sampling  & \textbf{20.846} &  \textbf{0.717}  & \textbf{0.205} \\
\Xhline{3\arrayrulewidth}
\end{tabular}
\end{adjustbox}
\end{table}

\subsubsection{Effectiveness of filter strategies}
As shown in Table \ref{table:Ablation_filter}, we present the effectiveness of the proposed filter strategies through various configurations. 
It can be seen that each filter strategy contributes to improved reconstruction quality, as shown in (b)-(d). 
Moreover, their combination can achieve better results, demonstrating their effectiveness in eliminating unreliable geometric constraints and providing informative supervision for sparse-input view synthesis.

\subsubsection{Effectiveness of sparse-friendly sampling}
In Table \ref{table:ablation_sampling}, we investigate the effectiveness of the non-max weight regularization and max weight densification operations in SFS. 
The experiments show that both operations enhance the representation of Gaussian ellipsoids, leading to improved PSNR.

\subsubsection{Number of looping}
To investigate the number of looping in the Progressive Gaussian Initialization (PGI) strategy, we present comparison results in Table \ref{table:ablation_loop}. 
Increasing the number of loops results in more initialization points, which can enhance rendering performance. However, excessive looping may lead to performance degradation. This is primarily because pseudo-views added at later stages may be distant from the training views, leading to inaccurate initialization points and ultimately impairing performance. Therefore, we selected three loops as the final setting for our experiments.

\subsubsection{Number of pseudo views}
To examine the impact of the number of pseudo views in the Progressive Gaussian Initialization (PGI) strategy, we conducted ablation experiments, as shown in Table \ref{table:ablation_pseudo_views}. 
The results indicate that adding 12 pseudo images at each Gaussian initialization, equivalent to four times the training views produces better results. Thus, we adopt this as the final configuration.

\begin{table}[!t]
\renewcommand{\arraystretch}{1.1}
\caption{Ablation study for the number of loop in PGI.} 
\label{table:ablation_loop}
\centering 
\begin{adjustbox} {width=.95\linewidth}
\begin{tabular}{c | c  p{1.25cm}<{\centering}p{1.25cm}<{\centering}p{1.25cm}<{\centering} p{1.25cm}<{\centering} }
\Xhline{3\arrayrulewidth}
{Loop Number} & Point Number & PSNR$\uparrow$ & SSIM$\uparrow$ & LPIPS$\downarrow$ \\ 
\hline
0   & 1921    & 20.203   & 0.677   & 0.225  \\
1   & 6370    & 20.773   & 0.711   & 0.209  \\
2   & 8531    & 20.773   & 0.714   & 0.207  \\
3   & 9903    & \textbf{20.846} &  \textbf{0.717}  & \textbf{0.205} \\
4   & 11078   & 20.717   & 0.715   & 0.206  \\
5   & 11922   & 20.761   & 0.716   & 0.205  \\ 
\Xhline{3\arrayrulewidth}
\end{tabular}
\end{adjustbox}
\end{table}

\begin{table}[!t]
\renewcommand{\arraystretch}{1.1}
\caption{Ablation study for the number of pseudo views.} 
\label{table:ablation_pseudo_views}
\centering 
\begin{adjustbox} {width=.9\linewidth}
\begin{tabular}{c | p{1.25cm}<{\centering}p{1.25cm}<{\centering}p{1.25cm}<{\centering} p{1.25cm}<{\centering} }
\Xhline{3\arrayrulewidth}
{Number of Pseudo Views}  & PSNR$\uparrow$ & SSIM$\uparrow$ & LPIPS$\downarrow$ \\ 
\hline
3       & 20.812   & 0.714   & 0.207  \\
6       & 20.838   & 0.715   & 0.205  \\
12      & \textbf{20.846} &  \textbf{0.717}  & 0.205 \\
24      & 20.764   & 0.717   & \textbf{0.204}  \\
48      & 20.814   & 0.716   & \textbf{0.204}  \\
\Xhline{3\arrayrulewidth}
\end{tabular}
\end{adjustbox}
\end{table}

\section{Concluding remarks}
\label{sec:Discussion}

This paper proposes LoopSparseGS, an innovative 3DGS-based framework for novel view synthesis using sparse input data.  
In LoopSparseGS, the proposed loop-based strategies, including Progressive Gaussian Initialization (PGI) and Depth-alignment Regularization (DAR), provide denser Gaussians and precise geometric information for effective scene coverage. 
Additionally, the proposed Sparse-friendly Sampling (SFS) strategy enhances Gaussian densification unique to sparse-input scenes, facilitating the generation of photo-realistic images.
Extensive experimental results demonstrate that our approach outperforms existing SOTA methods in sparse-input novel view synthesis across indoor, outdoor, and object-level scenes at various image resolutions.

{
    \small
    \bibliographystyle{ieeenat_fullname}
    \bibliography{main}
}


\end{document}